\documentclass[review]{elsarticle}

\usepackage[T1]{fontenc}
\usepackage{amssymb}
\usepackage{amsmath}
\usepackage[hidelinks]{hyperref}
\usepackage{float}
\usepackage{verbatim} 
\usepackage{graphicx,subfigure}
\usepackage{multirow}
\usepackage{multicol}
\usepackage{color}
\usepackage{apalike}
\restylefloat{figure}
\restylefloat{table}
\usepackage{xcolor,colortbl}
\usepackage{diagbox}
\usepackage{makecell}
\usepackage{lineno}

\usepackage{booktabs}
\usepackage[table]{xcolor}
\usepackage{makecell}
\usepackage{bm}
\usepackage{caption}


\journal{Expert Systems with Applications}

\biboptions{authoryear}

\begin{document}
\sloppy
\begin{frontmatter}

\begin{titlepage}
\begin{center}
\vspace*{1cm}


\textbf{\Large LQ-rPPG: A Label-Quantized Coarse-to-Fine Learning Framework for Remote Physiological Measurement}

\vspace{1.5cm}

Jun Seong Lee$^{a,b}$ (jslee0708@etri.re.kr), Samyeul Noh$^{a,b}$ (samuel@etri.re.kr), Changki Sung$^{b}$ (cs1032@kaist.ac.kr), Hyun Myung$^{b,*}$ (hmyung@kaist.ac.kr)\\

\hspace{10pt}

\begin{flushleft}
\small  
$^a$ Electronics and Telecommunications Research Institute, 218 Gajeong-ro, Yuseong-gu, Daejeon 34129, Republic of Korea \\
$^b$ School of Electrical Engineering, Korea Advanced Institute of Science and Technology, 291 Daehak-ro, Yuseong-gu, Daejeon 34141, Republic of Korea

\vspace{1cm}
\textbf{Declaration of Interest statement:}\\
The authors declare that they have no known competing financial interests or personal relationships that could have appeared to influence the work reported in this paper.


\vspace{1cm}
\textbf{Corresponding Author:} \\
Hyun Myung \\
School of Electrical Engineering, Korea Advanced Institute of Science and Technology, 291 Daehak-ro, Yuseong-gu, Daejeon 34141, Republic of Korea \\
Tel: +82-42-350-7651 \\
Email: hmyung@kaist.ac.kr

\end{flushleft}        
\end{center}
\end{titlepage}

\title{LQ-rPPG: A Label-Quantized Coarse-to-Fine Learning Framework for Remote Physiological Measurement}

\author[label1,label2]{Jun Seong Lee}
\ead{jslee0708@etri.re.kr}

\author[label1,label2]{Samyeul Noh}
\ead{samuel@etri.re.kr}

\author[label2]{Changki Sung}
\ead{cs1032@kaist.ac.kr}

\author[label2]{Hyun Myung \corref{cor1}}
\ead{hmyung@kaist.ac.kr}

\cortext[cor1]{Corresponding author.}
\address[label1]{Electronics and Telecommunications Research Institute, 218 Gajeong-ro, Yuseong-gu, Daejeon 34129, Korea}
\address[label2]{School of Electrical Engineering, Korea Advanced Institute of Science and Technology, 291 Daehak-ro, Yuseong-gu, Daejeon 34141, Korea}

\begin{abstract}
Remote photoplethysmography (rPPG) enables non-contact measurement of physiological signals from facial videos, offering strong potential for remote healthcare and daily health monitoring. Driven by this potential, various deep learning-based rPPG methods have been proposed to improve rPPG estimation. However, previous deep learning-based rPPG methods have paid little attention to the quality of training labels and their impact on model learning. Contact-based PPG signals used as training labels often contain noise and variability caused by motion artifacts, inconsistent sensor contact, and morphological distortions. Such label inconsistency can lead models to overfit to the label noise and variability and consequently degrade generalization performance. To address this issue, we propose LQ-rPPG, a label-quantized coarse-to-fine learning framework for robust rPPG estimation. LQ-rPPG consists of a label quantization module and a coarse-to-fine rPPG estimation model. The label quantization module transforms continuous PPG signals into multi-bit quantized pseudo labels with reduced noise and variability. The coarse-to-fine estimation model progressively refines rPPG signals under hierarchical supervision guided by the multi-bit pseudo labels. This design alleviates overfitting to label-specific variations and enables the model to learn structured and consistent representations. As a result, LQ-rPPG achieves robust and generalizable rPPG estimation even under challenging conditions. Experiments on multiple benchmark datasets demonstrate that LQ-rPPG achieves strong performance in both intra- and cross-dataset evaluations, while reducing parameters and multiply-accumulate operations by 88\% and 29\%, respectively, and increasing throughput by 191\%. The code is available at \url{https://github.com/Anonymous-repo-code/LQ-rPPG}.
\end{abstract}

   \end{frontmatter}


\section{Introduction}
\label{introduction}
Remote photoplethysmography (rPPG) is a non-contact optical technique that estimates blood volume pulse (BVP) signals from facial videos by analyzing subtle skin color variations induced by cardiac activity~\citep{verkruysse2008remote}. Specifically, by tracking pixel-level color changes over time, we can estimate BVP signals from facial videos, and the resulting video-based BVP signals are commonly referred to as rPPG signals. From rPPG signals, key physiological indicators such as heart rate (HR) and heart rate variability (HRV) can be derived, which in turn provide important information for assessing cardiovascular health and autonomic nervous system function~\citep{chen2018video, yu2021facial, zhang2023trusted, zhang2024temporal}. In contrast to traditional contact-based sensor methods that require attaching dedicated physiological sensors to the skin, rPPG enables fully non-contact and remote acquisition of physiological signals using only a standard camera. With these advantages, rPPG offers strong potential and broad applicability in remote healthcare and daily health monitoring~\citep{tohma2021evaluation, ma2025non, debnath2025comprehensive}.

Driven by the potential of rPPG, recent studies have explored deep learning-based methods for accurate and robust rPPG estimation. Early deep learning-based rPPG methods~\citep{vspetlik2018visual, chen2018deepphys} primarily used 2D convolutional neural networks (CNNs) to capture spatial features in facial regions, such as subtle cardiac-induced skin color changes. Subsequent rPPG methods enhanced these spatial approaches by incorporating temporal modeling to capture the dynamic characteristics of rPPG signals. Such efforts include combining CNNs with recurrent neural networks (RNNs)~\citep{lee2020meta, nowara2021benefit}, constructing spatiotemporal maps (STMaps)~\citep{niu2019rhythmnet, niu2020video}, and utilizing 3D CNNs~\citep{yu2019remote, botina2022rtrppg, lee2023lstc, li2023learning} to jointly capture spatial and temporal dynamics. Following these advances, more recent rPPG methods introduced Transformer-based and Mamba-based architectures to effectively model global context, including the quasi-periodic nature of rPPG signals. Transformer-based methods~\citep{yu2022physformer, yu2023physformer++, shao2023tranphys, choi2024fusion, zou2025rhythmformer} used self-attention mechanisms to model long-range dependencies, while Mamba-based methods~\citep{luo2024physmamba, zou2025rhythmmamba} utilized selective state-space models to achieve comparable long-range modeling with higher computational efficiency. Collectively, these architectural and methodological advances have substantially improved the accuracy and robustness of rPPG estimation. However, prior rPPG methods have mainly focused on improving rPPG performance by designing powerful backbones or spatiotemporal modeling schemes, while comparatively underexploring how the quality and structure of the supervision signals themselves affect model robustness and generalization.

\begin{figure}
\centering
\includegraphics[width=0.98\linewidth]{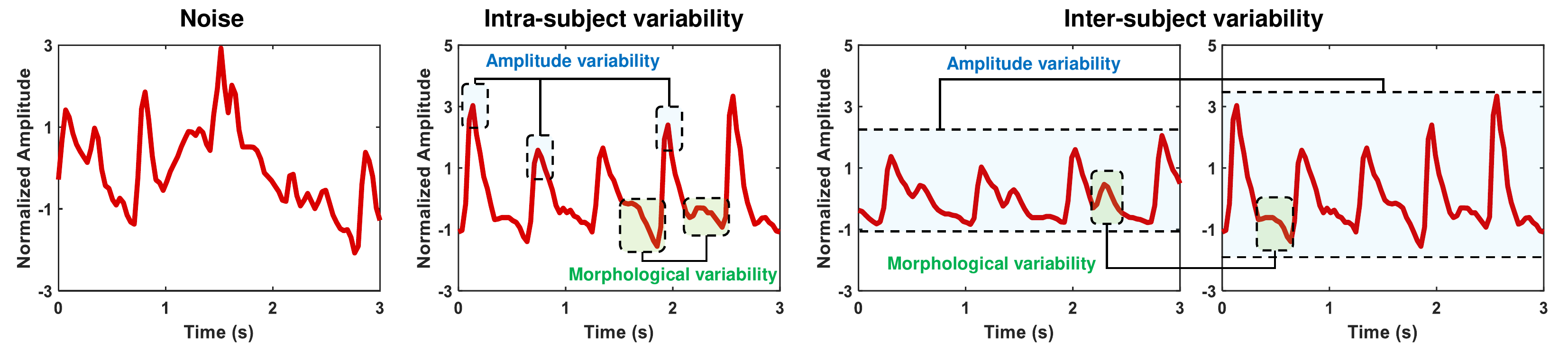}
\caption{Examples of noise and variability in contact-based PPG signals used as ground truth. From left to right: (1) Noise --- signal degradation caused by artifacts such as motion or sensor interference; (2) Intra-subject variability --- morphological and amplitude variations occurring within the same subject; and (3) Inter-subject variability --- differences in overall amplitude and waveform shape observed across subjects. All PPG signals were normalized using z-score normalization, consistent with the preprocessing used during model training.}
\label{fig:variability}
\end{figure}

In practice, contact-based PPG signals used as ground truth for rPPG model training are often far from ideal, containing various forms of noise and variability, such as motion-induced artifacts, inconsistent sensor contact, morphological distortions, and amplitude fluctuations (see Fig.~\ref{fig:variability} and \ref{appendix_quant_label_quality}). Even for the same subject and recording setup, the corresponding PPG signals can differ in waveform morphology, amplitude, and noise level, which weakens the consistency of supervision and makes it difficult for models to learn stable and generalizable physiological representations. Moreover, because rPPG estimation inherently relies on subtle physiological cues and the available training data in the rPPG domain are typically limited, models trained with such noisy and inconsistent labels are especially prone to overfitting to subject- or environment-specific variations rather than learning core physiological patterns, which ultimately degrades generalization performance.

One promising direction to alleviate these challenges is to adopt a label-discretization-based supervision strategy. In recent deep learning studies, such discretization-based learning has attracted attention as a paradigm for improving stability and robustness under uncertain label conditions, and its effectiveness has been demonstrated in various computer vision tasks, including depth estimation~\citep{bhat2021adabins, bhat2022localbins, shao2023iebins} and human pose estimation~\citep{li2022simcc}. Although the noise and variability in PPG signals are not strictly equivalent to label uncertainty, they are similar in that they disrupt the consistency of supervision and hinder stable model learning. From this perspective, a discretization-based supervision strategy that transforms continuous PPG signals into structured discrete labels can serve as a promising alternative for learning more stable and generalizable rPPG representations under noisy and variable supervisory signals.

However, discretization-based supervision has been relatively underexplored in the rPPG literature, and the discretization schemes commonly adopted in other domains~\citep{bhat2021adabins, bhat2022localbins, li2022simcc}, which map continuous targets at a single resolution and use them as supervision, are not well suited to rPPG learning. Such single-resolution discretization provides limited flexibility in simultaneously addressing two competing requirements: suppressing noise and variability in contact-based PPG signals while preserving fine-grained cardiovascular waveform characteristics. If the discretization is too coarse, it fails to capture subtle waveform structures and amplitude variations of cardiovascular rhythms, leading to limited representational capacity. Conversely, overly fine discretization becomes highly sensitive to label noise and variability, undermining the original goal of improving robustness. Consequently, supervision that relies solely on a single discrete label space exposes an inherent trade-off between stability and waveform fidelity, making it difficult to achieve both stable and high-fidelity rPPG estimation.

To address these challenges, we propose LQ-rPPG, a label-quantized coarse-to-fine learning framework for robust rPPG estimation. LQ-rPPG operationalizes the label discretization paradigm through multi-level quantization of PPG signals and hierarchical coarse-to-fine supervision, thereby mitigating the adverse effects of label noise and variability, and alleviating the stability--fidelity trade-offs involved in the label discretization process. Concretely, the framework consists of two main components: a label quantization (LQ) module and a coarse-to-fine (C2F) estimation model. The LQ module converts continuous PPG signals into multi-bit quantized pseudo labels with reduced noise and variability. These pseudo labels exhibit different levels of stability and fidelity depending on their bit resolution, which enables hierarchical supervision. Building on these quantized pseudo labels, the C2F model estimates rPPG signals in a hierarchical manner by first capturing coarse global physiological rhythms under low-bit-resolution supervision (e.g., using 1- or 2-bit quantized pseudo labels) and then refining fine-grained waveform characteristics under high-bit-resolution supervision. Through this structured and hierarchical learning process, LQ-rPPG effectively mitigates the label noise and variability. As a result, it achieves robust and generalizable rPPG estimation across diverse conditions.

The main contributions of this study are summarized as follows:
\begin{enumerate}
\item We introduce a label-centric perspective, relatively underexplored in prior rPPG studies, emphasizing the impact of label noise and variability on model robustness and generalization performance; while recent Transformer- and Mamba-based rPPG methods mainly focus on strengthening backbones and temporal modeling, our work instead focuses on supervision quality and structure.
\item We propose LQ-rPPG, a novel framework that quantizes PPG signals into multi-bit pseudo labels and performs coarse-to-fine estimation under hierarchical supervision for robust and efficient rPPG estimation.
\item Extensive experiments demonstrate that LQ-rPPG achieves strong performance in both intra- and cross-dataset scenarios.
\item LQ-rPPG also achieves strong computational efficiency, maintaining competitive performance while reducing the number of parameters and multiply-accumulate operations (MACs) by 88\% and 29\%, respectively, and increasing throughput by 191\%.
\end{enumerate}

The remainder of this paper is organized as follows. Section~\ref{sec2} reviews the related work. Section~\ref{sec3} presents the details of the proposed LQ-rPPG framework. Section~\ref{sec4} reports the experimental results and discussions. Finally, Section~\ref{sec5} concludes the paper.

\section{Related work}
\label{sec2}
\subsection{Traditional signal processing-based rPPG methods}
\citet{verkruysse2008remote} first demonstrated the presence of subtle pulsatile signals in facial videos that reflect cardiac activity, which inspired a range of traditional rPPG methods based on signal processing techniques. These signal processing-based methods can be broadly categorized into region of interest (ROI)-based, source separation-based, and color transformation-based methods.

ROI-based methods~\citep{sun2011motion, li2014remote, feng2015dynamic, bobbia2016remote} focus on selecting stable skin regions to improve the quality of rPPG signals. For example, \citet{li2014remote} used facial landmarks, and \citet{bobbia2016remote} employed skin segmentation to track stable skin regions.

Source separation-based methods~\citep{poh2010advancements, poh2010noncontact, mcduff2014improvements, wang2014exploiting} aim to extract pulse-related signals by decomposing RGB color signals into physiological and non-physiological components. For instance, \citet{poh2010advancements} applied independent component analysis (ICA), while \citet{wang2014exploiting} used principal component analysis (PCA) to decompose color signals into statistically independent or uncorrelated sources.

Color transformation-based methods~\citep{dehaan2013chrom, dehaan2014pbv, wang2016algorithm} perform linear transformations of RGB signals, such as weighted channel combinations or subspace projections, to enhance the pulsatile information embedded in facial color changes. Specifically, \citet{dehaan2013chrom} proposed the CHROM method using chrominance differences, and \citet{wang2016algorithm} introduced the POS method that projects color signals onto a plane orthogonal to the skin tone.

Despite their simplicity and efficiency, traditional signal processing-based methods suffer from limited performance under real-world conditions involving motion, illumination, and skin tone variations.

\subsection{Deep learning-based rPPG methods}
To address the limitations of traditional signal processing-based approaches, researchers have recently explored deep learning-based rPPG methods. Instead of relying on manually designed filters or feature extraction pipelines, these methods learn nonlinear spatial and temporal representations directly from data, thereby improving robustness to motion, illumination, and skin tone variations.

Early deep learning-based methods~\citep{tang2018noncontact, vspetlik2018visual, chen2018deepphys} were based on 2D CNNs. By utilizing the nonlinear representation learning capability of 2D CNNs, these methods learned spatial representations of subtle facial color variations. For instance, DeepPhys~\citep{chen2018deepphys} introduced an attention-enhanced 2D CNN framework that adaptively focused on skin regions and extracted color variation features informative of rPPG signals. However, these early approaches largely emphasize frame-level cues, which limits their ability to capture the sequential and rhythmic temporal dynamics inherent to rPPG signals.

Motivated by this limitation, subsequent works introduced spatiotemporal architectures that explicitly model temporal dynamics across video frames \citep{lee2020meta, nowara2021benefit, niu2019rhythmnet, niu2020video, yu2019remote, botina2022rtrppg, lee2023lstc, li2023learning}. Among them, Meta-rPPG~\citep{lee2020meta} combined CNNs with RNNs to capture temporal dependencies, whereas RhythmNet~\citep{niu2019rhythmnet} utilized STMaps to jointly encode spatial and temporal cues. PhysNet~\citep{yu2019remote} adopted 3D CNNs to directly learn spatiotemporal representations from facial videos. While these designs improve temporal modeling beyond frame-level cues, they still mainly capture local or short-range dynamics, which limits their ability to fully model the quasi-periodic structure of rPPG signals, particularly under challenging scenarios.

More recently, advanced architectures have been proposed to capture global contextual relationships and long-range temporal dependencies in facial videos. Transformer-based methods~\citep{yu2022physformer, yu2023physformer++, choi2024fusion, zou2025rhythmformer, liu2025style, li2025lst}, such as PhysFormer~\citep{yu2022physformer}, leverage self-attention mechanisms to model long-range dependencies and enhance global contextual representation learning. However, self-attention typically incurs substantial computational and memory overhead for long video sequences, which can limit practical efficiency. To address this efficiency concern, Mamba-based methods~\citep{luo2024physmamba, zou2025rhythmmamba}, including RhythmMamba~\citep{zou2025rhythmmamba}, employ selective state-space models to efficiently model long-term temporal dependencies with reduced computational complexity.

Overall, prior deep learning-based rPPG methods have mainly progressed by refining architectures and representations for feature extraction. In contrast, the quality and consistency of supervisory signals have received relatively little attention, even though label noise and variability in contact-based PPG supervision can hinder learning and degrade generalization. This gap motivates our work, which shifts the emphasis from improving feature representations to a label-centric perspective for more structured and generalizable rPPG learning.

\subsection{Discretized supervision in deep learning}
Discretized supervision has recently emerged as a practical strategy in deep learning to enhance training stability and robustness under uncertain or noisy label conditions. By partitioning continuous label values into discrete intervals, this approach regularizes the label space and reduces the influence of local label fluctuations or ambiguities during optimization. Such discretization provides more reliable supervisory signals, guiding models to learn representations that are less sensitive to label noise and more consistent across varying data conditions.

This paradigm has been successfully applied across various computer vision tasks, where continuous supervision is frequently noisy, unstable, or inherently ambiguous. In depth estimation, continuous depth labels can be corrupted by sensor noise and observation uncertainty, and they can also exhibit discontinuities near object boundaries. To mitigate these issues, discretization-based formulations that represent depth using bins or distributions have been explored. For instance, AdaBins~\citep{bhat2021adabins} introduced adaptive binning to more effectively represent depth distributions. Extending this idea, LocalBins~\citep{bhat2022localbins} and IEBins~\citep{shao2023iebins} refined the bin structures through locally adaptive and progressively optimized strategies. In human pose estimation, ground-truth joint coordinates can become ambiguous under occlusion or low resolution; accordingly, discretization strategies that reformulate continuous coordinate regression into classification-based formulations have been adopted. SimCC~\citep{li2022simcc} reformulated coordinate regression as one-dimensional classification along each axis, enhancing localization precision under ambiguity. In oriented object detection, continuous angle regression can be unstable due to the circular nature of angles and directional ambiguity. \citet{yang2020arbitrary} mitigated these issues by discretizing angular predictions to address circularity and reduce directional uncertainty. Overall, these discretization-based approaches share a common motivation: they structure continuous supervision to alleviate noise, variability, and ambiguity, thereby reducing the learning burden of fitting unstable label fluctuations.

The demonstrated effectiveness of discretization strategies in such noisy or ambiguous continuous-label settings suggests that this concept can be extended to the rPPG domain. In rPPG learning, contact-based PPG signals often exhibit substantial variability arising from motion artifacts, inconsistent sensor contact, and physiological differences across subjects. Although such variability is not equivalent to label uncertainty in a strict sense, it similarly disrupts the consistency of supervision and can be regarded as a form of supervision uncertainty that is conceptually analogous to the noisy, unstable, or inherently ambiguous continuous-label scenarios discussed above. Based on these observations, this study proposes a discretized supervision approach tailored to rPPG that quantizes PPG signals and leverages coarse-to-fine learning to improve robustness and generalization in rPPG estimation.

\section{Label-Quantized Coarse-to-Fine Learning Framework}
\label{sec3}
\subsection{Overview of LQ-rPPG}
One effective way to alleviate the impact of label noise and variability in the rPPG task is to quantize PPG signals and reformulate the rPPG task as a classification-based learning problem.
Quantization compresses the label space and reduces variability, leading to more structured and generalizable learning.
However, the quantization process inherently involves a trade-off between variability reduction and physiological fidelity.
For example, low-bit labels suppress subtle fluctuations and emphasize global physiological rhythms, providing simple and consistent supervision but losing fine-grained details for precise estimation.
In contrast, high-bit labels preserve detailed amplitude and waveform details that are crucial for capturing fine-grained physiological dynamics but are more sensitive to noise and variability.

To alleviate the impact of label noise and variability while addressing the trade-off introduced by quantization, we propose LQ-rPPG, which quantizes PPG signals into multiple bit levels and employs a hierarchical coarse-to-fine learning strategy.
The framework first learns stable global patterns under low-bit supervision and then progressively refines representations with high-bit pseudo labels.
This hierarchical process effectively integrates the stability of low-bit supervision with the precision of high-bit guidance, leading to robust and accurate rPPG estimation.

\begin{figure}[t]
\centering
\includegraphics[width=0.95\linewidth]{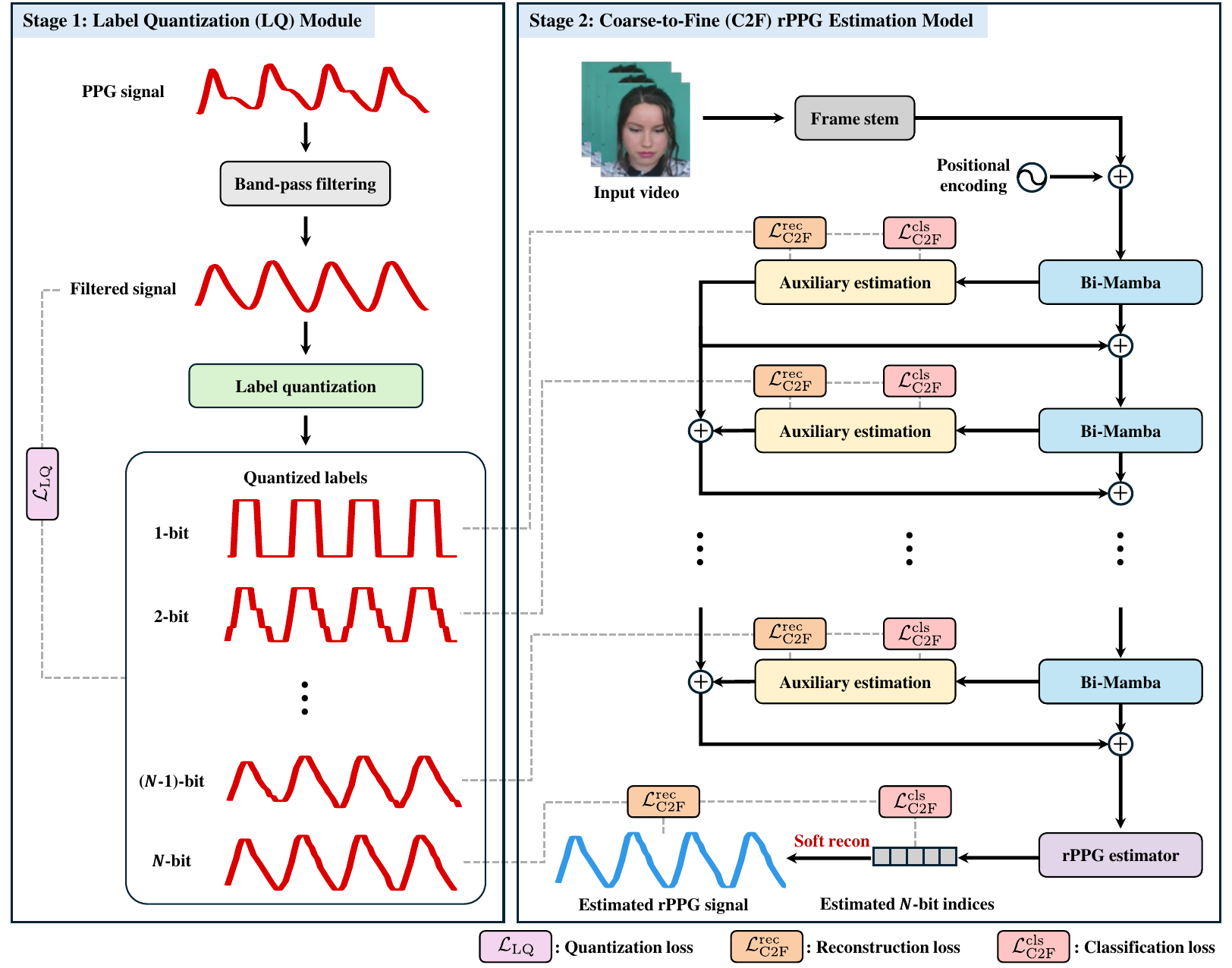}
\caption{Overview of LQ-rPPG. The framework consists of (a) a label quantization module (Stage~1) that generates multi-bit pseudo labels, and (b) a coarse-to-fine rPPG estimation model (Stage~2) that progressively estimates rPPG signals under hierarchical supervision.}
\label{fig:overall}
\end{figure}

The overall architecture of LQ-rPPG is illustrated in Fig.~\ref{fig:overall}.
The proposed LQ-rPPG takes facial videos as input and estimates continuous rPPG signals.
It consists of two main stages: (1) label quantization, which converts PPG signals into multi-bit quantized pseudo labels, and (2) coarse-to-fine rPPG estimation, which predicts rPPG signals under hierarchical supervision guided by the pseudo labels.
These two stages are structurally coupled through the multi-bit quantized pseudo labels generated in Stage~1 and used to guide the hierarchical estimation process in Stage~2.

In the first stage, a PPG signal is initially band-pass filtered to suppress noise and retain the physiological components within the cardiac frequency band. The filtered signal is then processed through a learning-based label quantization process, which discretizes the signal multiple times with different bit levels to generate multi-level quantized pseudo labels.
The quantization process is guided by a quantization loss ($\mathcal{L}_{\text{LQ}}$), which enforces temporal and frequency consistency between the input and the quantized output. Through the process, the continuous label space is transformed into a compact discrete label space with reduced variability and preserved physiologically meaningful rhythms. The resulting pseudo labels, each corresponding to a specific bit level, exhibit varying degrees of variability reduction and physiological fidelity and are used as hierarchical guidance in the subsequent coarse-to-fine learning stage.

In the second stage, a video input is transformed into compact temporal features through the frame stem module~\citep{zou2025rhythmmamba}, and positional encoding is added to emphasize temporal characteristics.
The resulting features are progressively refined through a sequence of Bi-Mamba~\citep{zhu2024vision} blocks, each paired with an auxiliary estimation branch guided by the corresponding bit-level pseudo labels generated in Stage~1.
Each Bi-Mamba block captures temporal dependencies from the input representations.
The auxiliary branch predicts bit-level labels through classification and soft reconstruction based on a classification loss ($\mathcal{L}^{\text{cls}}_{\text{C2F}}$) and a reconstruction loss ($\mathcal{L}^{\text{rec}}_{\text{C2F}}$). The outputs from the auxiliary branches are cumulatively aggregated and passed to the subsequent Bi-Mamba blocks.
Through the process, the feature representations are progressively refined from coarse to fine in a hierarchical manner.
At the final refinement step, the refined representation integrates the output of the last Bi-Mamba block with all preceding bit-level estimations (from 1 to ($N$-1) bits), producing a rich temporal representation that combines coarse global rhythms with fine local variations of physiological signals.
Finally, an rPPG estimator predicts an rPPG signal from the refined representation under the guidance of the $N$-bit pseudo label.

Overall, the proposed framework adopts a two-stage learning strategy. In Stage 1, noisy and variable continuous PPG signals used as training labels are converted into multi-bit quantized pseudo labels, which have two key implications. First, compressing the continuous label space into a discrete space mitigates sample-specific waveform and amplitude fluctuations, thereby enabling more stable and consistent supervision. Second, pseudo labels at different bit levels have different supervision granularities, enabling coarse-to-fine learning in Stage 2: low-bit pseudo labels guide the model to focus on coarse patterns such as overall cardiac rhythms, whereas high-bit pseudo labels guide it to incorporate fine-grained waveform and amplitude variations. In Stage 2, the rPPG signal is estimated from video inputs under the hierarchical supervision of these multi-bit pseudo labels. Specifically, the model first learns rhythm-oriented coarse features that remain relatively consistent across diverse conditions under low-bit supervision, and then progressively refines the rPPG estimation by incorporating high-bit supervision. As a result, this two-stage design reduces overfitting to label variability while maintaining precision, leading to robust and generalizable rPPG estimation.

The following sections provide further details: Section~\ref{sec:LQ} introduces the label quantization module, Section~\ref{sec:Opt_LQ} presents the optimization objective for the label quantization module, Section~\ref{sec:C2F} describes the coarse-to-fine rPPG estimation model, and Section~\ref{sec:Opt_C2F} defines the optimization objective for the coarse-to-fine rPPG estimation model.

\paragraph{Terminology}
In this paper, we use \textit{pseudo labels} as an umbrella term for the supervisory signals generated in Stage~1 from contact-based PPG signals through our label quantization process.
To emphasize the bit-level quantization, we refer to them as \textit{quantized pseudo labels}.
The term \textit{discretization} (or \textit{discretize}) is used in its general sense, i.e., mapping a continuous signal to a finite set of values, and is not used as a dedicated label name.

\subsection{Label quantization module}
\label{sec:LQ}

\begin{figure}[t]
\centering
\includegraphics[width=0.98\linewidth]{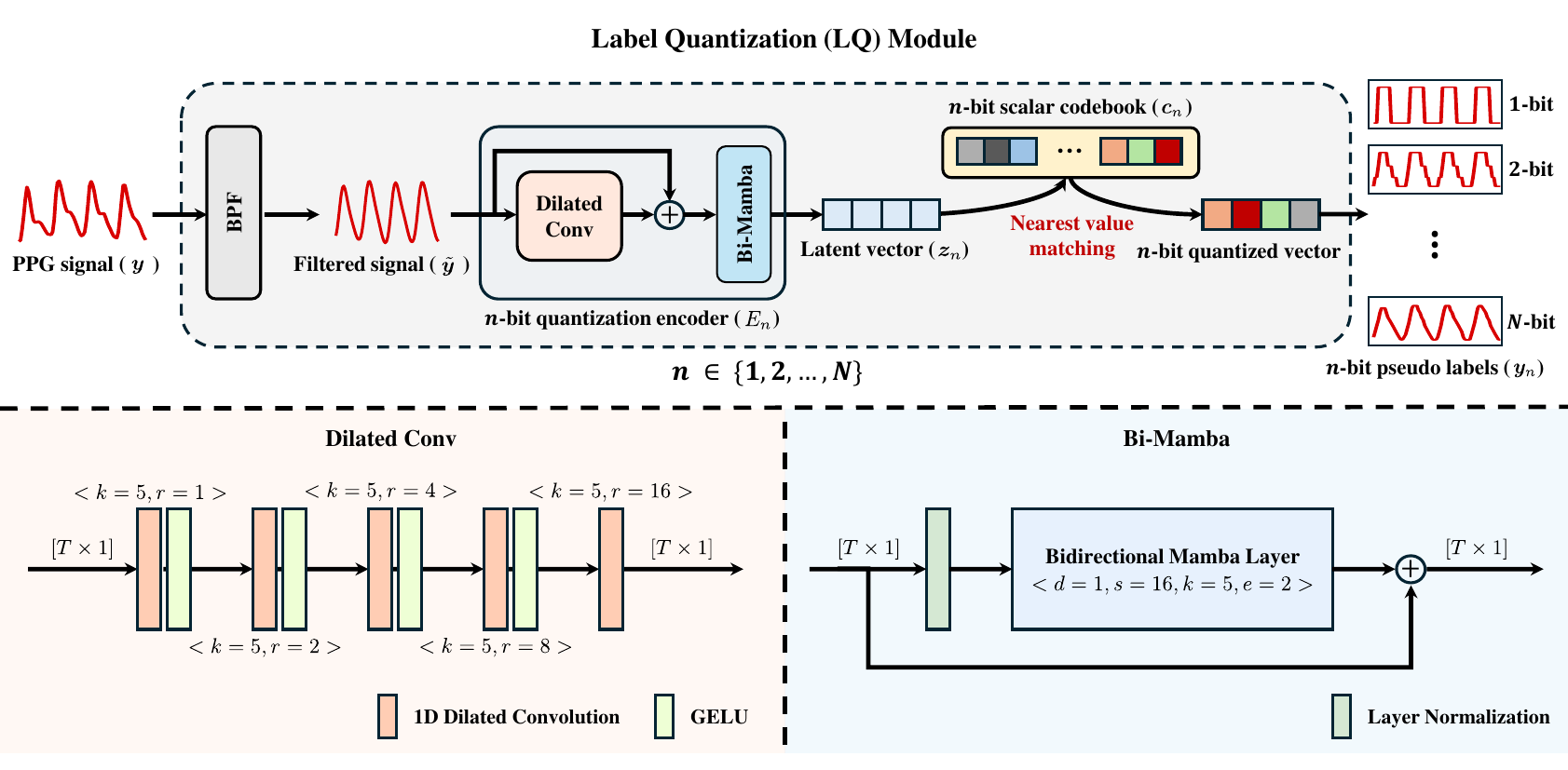}
\caption{
Label quantization module. 
The module transforms a continuous PPG signal into multi-bit quantized pseudo labels with reduced noise and variability. 
In the diagram, square brackets $[\;]$ denote the dimensionality of data, and angle brackets $\langle\;\rangle$ indicate layer-specific hyperparameters. 
In the dilated convolution layers, $k$ and $r$ refer to the kernel size and dilation rate, respectively. 
In the bidirectional Mamba layer, $d$, $s$, $k$, and $e$ denote the model dimension, state dimension, convolution kernel size, and expansion ratio, respectively.
}
\label{fig:LQ}
\end{figure}

In the rPPG task, target PPG signals are often corrupted by motion noise and exhibit intra-subject, inter-subject, and cross-domain variability.
Using such noisy and unstable signals as regression targets can cause overfitting to sample-specific artifacts and degrade model performance.

To overcome this issue, we propose a label quantization (LQ) module.
The LQ module generates quantized labels with reduced noise and variability, which allows the rPPG task to be reformulated as a classification-based learning problem.
This discretized formulation regularizes the label space and enables an rPPG estimation model to learn more structured and generalized representations under stable supervision.
As a result, the overall learning process becomes more robust and effective, leading to improved generalization across diverse conditions.

Figure~\ref{fig:LQ} illustrates the proposed LQ module. Given a PPG signal $\bm{y} \in \mathbb{R}^T$, where $T$ is the temporal length, the module generates multiple levels of pseudo labels $\bm{y}_n$ by applying quantization with different bit depths $n \in \{1, 2, \cdots, N\}$, where $N$ denotes the maximum bit depth.

For each $n$, $\bm{y}$ is first preprocessed by a band-pass filter to reduce noise and preserve the main cardiac frequency components, following the frequency range (0.75--2.5 Hz) adopted in a previous rPPG study~\citep{liu2023rppg}. The filtered signal $\tilde{\bm{y}}$ is then fed into an $n$-bit quantization encoder $E_n$, producing a latent vector $\bm{z}_n \in \mathbb{R}^T$. 
$E_n$ consists of a dilated convolution block and a Bi-Mamba block~\citep{zhu2024vision}.
The dilated convolution block models local temporal characteristics, while the Bi-Mamba block captures global dependencies.
The adoption of the Mamba-based architecture is motivated by prior evidence that it effectively captures long-range dependencies with high computational efficiency~\citep{luo2024physmamba, zou2025rhythmmamba}.

The dilated convolution block is composed of five 1D dilated convolution layers with a kernel size of $5$ and dilation rates of 1, 2, 4, 8, and 16.
Gaussian Error Linear Unit (GELU) activation is applied after each layer except the last. 
Such a stacked dilated design yields an effective receptive field of 125 frames. 
Assuming a sampling rate of 30 Hz, the receptive field corresponds to approximately 4.2 seconds of temporal context, which covers at least two cardiac cycles even for subjects with low heart rates. 
Such a receptive field allows the block to model local temporal patterns. 

The Bi-Mamba block consists of a bidirectional Mamba layer and layer normalization, followed by a residual connection. 
The bidirectional Mamba layer uses the following parameters: model dimension $d=1$, state dimension $s=16$, kernel size $k=5$, and expansion ratio $e=2$. 
The layer models bidirectional and global temporal dependencies from the input sequence, while the residual connection facilitates gradient propagation and stabilizes the optimization process.

Consequently, $\bm{z}_n$ encodes both short- and long-range dependencies that vary with the quantization level $n$, which provides a temporally enriched representation for subsequent quantization. 
Formally, it is defined as:
\begin{equation}
\bm{z}_n = [z_{1,n}, z_{2,n}, \cdots, z_{T,n}]^\top = E_n(\tilde{\bm{y}}),
\end{equation}
where $z_{t,n}$ denotes the $t$-th latent element for the $n$-bit quantization, with $t = 1, 2, \cdots, T$.

To quantize $\bm{z}_n$, we define a learnable scalar codebook as:
\begin{equation}
\bm{c}_n = [c_{1,n}, c_{2,n}, \cdots, c_{2^n,n}]^\top \in \mathbb{R}^{2^n},
\end{equation}
where each $c_{k,n} \in \mathbb{R}$ denotes the $k$-th code in the $n$-bit quantization. For each latent element $z_{t,n}$, the nearest code index is obtained by:
\begin{equation}
i(z_{t,n}) = \operatorname*{argmin}_{k \in \{1, 2, \cdots, 2^n\}} \left\| z_{t,n} - c_{k,n} \right\|_2,
\end{equation}
and the corresponding index vector is formed as:
\begin{equation}
\bm{i}_n = [i(z_{1,n}), i(z_{2,n}), \cdots, i(z_{T,n})]^\top.
\end{equation}

Based on the index vector $\bm{i}_n$, each latent element $z_{t,n}$ is replaced by the corresponding code as $y_{t,n} = c_{i(z_{t,n}),n}$ for $t = 1, 2, \cdots, T$.
The final pseudo label is then obtained as:
\begin{equation}
\bm{y}_n = [y_{1,n}, y_{2,n}, \cdots, y_{T,n}]^\top.
\end{equation}

\subsection{Optimization objective for label quantization module}
\label{sec:Opt_LQ}

The LQ module aims to generate physiologically meaningful quantized pseudo labels that preserve core pulse-related information while reducing label noise and variability.
To achieve this, two loss components are used: a reconstruction loss $\mathcal{L}^{\text{rec}}_{\text{LQ}}$ and a feature commitment loss $\mathcal{L}^{\text{feat}}_{\text{LQ}}$.

The reconstruction loss $\mathcal{L}^{\text{rec}}_{\text{LQ}}$ adopts the formulation introduced in previous rPPG studies~\citep{zou2025rhythmformer, zou2025rhythmmamba}, where the loss function was designed to preserve physiological fidelity by enforcing temporal and spectral consistency between raw PPG and estimated rPPG signals. In the proposed LQ module, this principle is extended to ensure that the quantized pseudo label $\bm{y}_n$ remains temporally and spectrally aligned with the band-pass filtered signal $\tilde{\bm{y}}$. Although band-pass filtering can sacrifice fine-grained waveform details, the band-pass filtered signal is used as a reference to guide pseudo-label generation, reducing the influence of components outside the HR band and encouraging the pseudo labels to focus on the dominant heart-rate rhythm. Formally, the reconstruction loss is defined as:
\begin{align}
\mathcal{L}^{\text{rec}}_{\text{LQ}} 
= \lambda_{\text{time}} \cdot \mathrm{Neg}(\tilde{\bm{y}}, \bm{y}_n)
+ \lambda_{\text{freq}} \cdot \mathrm{CE}\big(\mathrm{maxIndex}(\mathrm{PSD}(\tilde{\bm{y}})), \mathrm{PSD}(\bm{y}_n)\big),
\label{eq:rec_lq}
\end{align}
where $\mathrm{Neg}(\cdot)$ denotes the negative Pearson correlation, $\mathrm{CE}(\cdot)$ represents the cross-entropy loss, and $\mathrm{PSD}(\cdot)$ refers to the power spectral density. The first term enforces temporal coherence between $\tilde{\bm{y}}$ and $\bm{y}_n$, and the second promotes spectral alignment by matching their dominant frequency components. Together, these terms ensure that the quantized pseudo labels remain temporally and spectrally consistent with the physiological signal.
  
The feature commitment loss $\mathcal{L}^{\text{feat}}_{\text{LQ}}$ encourages the encoder’s latent feature $\bm{z}_n$ to align with its corresponding quantized output $\bm{y}_n$, thereby stabilizing the quantization process and promoting codebook utilization:
\begin{equation}
\mathcal{L}^{\text{feat}}_{\text{LQ}} 
= \lambda_{\text{feat}} \cdot \left\| \bm{z}_n - \text{sg}(\bm{y}_n) \right\|_2^2,
\end{equation}
where $\text{sg}(\cdot)$ denotes the stop-gradient operator that prevents gradient propagation through the quantized label $\bm{y}_n$, allowing only the encoder $E_n$ to be updated. 
The codebook $\bm{c}_n$ is updated using an exponential moving average (EMA)~\citep{van2017neural} during training to ensure stable convergence.

The overall training objective is given by:
\begin{equation}
\mathcal{L}_{\text{LQ}} = 
\mathcal{L}^{\text{rec}}_{\text{LQ}} +
\mathcal{L}^{\text{feat}}_{\text{LQ}}.
\end{equation}
This objective enables the LQ module to produce physiologically meaningful pseudo labels with reduced variability and preserved physiological relevance. The relative contributions of each loss term are controlled by the coefficients $\lambda_{\text{time}}$, $\lambda_{\text{freq}}$, and $\lambda_{\text{feat}}$.  

\subsection{Coarse-to-fine rPPG estimation model}
\label{sec:C2F}
In the proposed framework, the LQ module generates multi-bit pseudo labels with different granularity and fidelity.
Low-bit pseudo labels exhibit reduced variability and emphasize global physiological rhythms, providing stable and simplified supervision.
In contrast, high-bit pseudo labels preserve fine waveform details, enabling precise supervision.
To effectively leverage these different levels of granularity and fidelity, we propose a coarse-to-fine (C2F) rPPG estimation model.
The C2F model first learns global cardiac rhythms from low-bit labels and then refines representations using high-bit labels.
Through this hierarchical learning process, the C2F model achieves robust and precise rPPG estimation.

\begin{figure}[t!]
\centering
\includegraphics[width=0.98\linewidth]{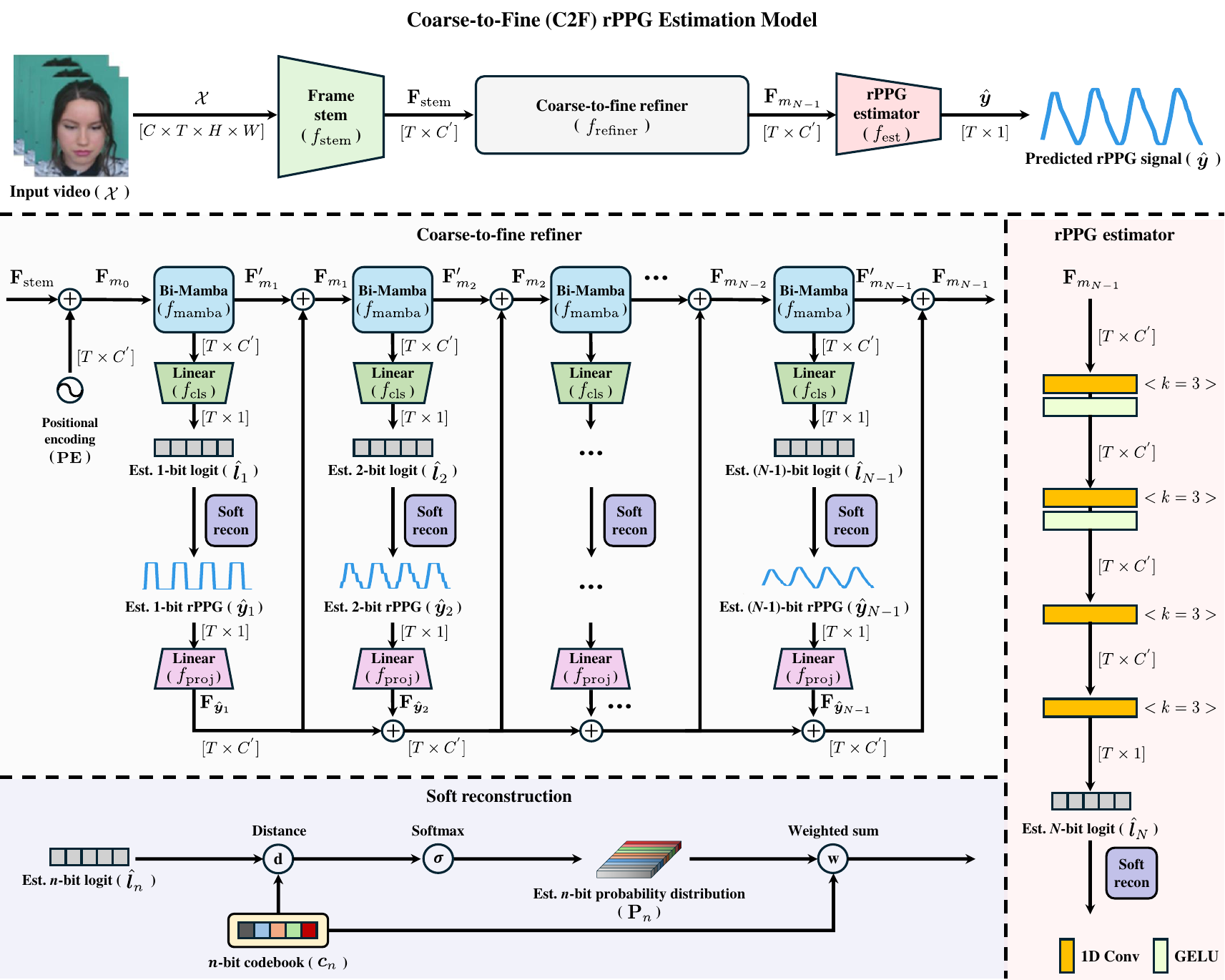}
\caption{
Coarse-to-fine rPPG estimation model. 
The model estimates rPPG signals from facial videos under hierarchical guidance by the multi-bit pseudo labels generated by Stage 1. 
The frame stem compresses input videos into sequential features, the coarse-to-fine refiner progressively refines temporal representations, and the rPPG estimator predicts the final rPPG signal. 
In the diagram, square brackets $[\;]$ denote the dimensionality of data, and angle brackets $\langle\;\rangle$ indicate the hyperparameters of each layer. 
In the convolutional layers, $k$ refers to the kernel size.
}
\label{fig:C2F}
\end{figure}

Figure~\ref{fig:C2F} shows the proposed C2F model. Given an input video $\mathcal{X} \in \mathbb{R}^{C \times T \times H \times W}$, 
where $C$, $T$, $H$, and $W$ denote the number of channels, temporal length, height, and width, respectively, 
the proposed model predicts an rPPG signal $\hat{\bm{y}} \in \mathbb{R}^{T}$.

The proposed C2F model consists of three modules: (i) a frame stem, (ii) a coarse-to-fine refiner, and (iii) an rPPG estimator.
The input video $\mathcal{X}$ is first processed by the frame stem module $f_{\text{stem}}(\cdot)$, following the design of~\citet{zou2025rhythmmamba}.
The frame stem compresses spatial information into temporal representations through global average pooling, transforming the input from $(C, T, H, W)$ to $(T, C')$ for treating rPPG as a time-series modeling problem. 
The resulting stemmed features are defined as:
\begin{equation}
\mathbf{F}_{\text{stem}} = f_{\text{stem}}(\mathcal{X}) \in \mathbb{R}^{T \times C'}.
\end{equation}

The stemmed features $\mathbf{F}_{\text{stem}}$ then enter the coarse-to-fine refiner $f_{\text{refiner}}(\cdot)$, which produces the last refined features:
\begin{equation}
\mathbf{F}_{m_{N-1}} = f_{\text{refiner}}(\mathbf{F}_{\text{stem}}) \in \mathbb{R}^{T \times C'}.
\end{equation}
Concretely, the refiner begins by adding learnable positional encodings $\mathbf{PE}$ to $\mathbf{F}_{\text{stem}}$ to emphasize the sequential structure, resulting in:
\begin{equation}
\mathbf{F}_{m_0} = \mathbf{F}_{\text{stem}} + \mathbf{PE} \in \mathbb{R}^{T \times C'}.
\end{equation}
Next, the encoded features $\mathbf{F}_{m_0}$ are refined through $N-1$ progressively finer steps, each indexed by the quantization bit depth $n \in \{1, 2, \cdots, N-1\}$. Each step strengthens the temporal structure under hierarchical supervision and includes three components: a Bi-Mamba block, a linear layer for classification, and another linear layer for feature projection. The Bi-Mamba block follows the same architecture as that used in the LQ module, with the model dimension increased to $d=64$ while keeping the other parameters fixed ($s=16$, $k=5$, $e=2$).
Both linear layers are implemented as fully connected layers, where the classification layer maps the feature dimension from $C'$ to 1, and the feature projection layer inversely maps from 1 back to $C'$.

At each step $n$, the Bi-Mamba block processes its input features to produce temporally refined features, denoted as $\mathbf{F}'_{m_{n}}$:
\begin{align}
\mathbf{F}'_{m_{n}} &= f_{\text{mamba}}^n\left(\mathbf{F}_{m_{n-1}}\right) \in \mathbb{R}^{T \times C'},
\end{align}
where $f_{\text{mamba}}^n(\cdot)$ denotes the Bi-Mamba block at step $n$.
The classification layer then takes $\mathbf{F}'_{m_{n}}$ as input and outputs the bit-level logits $\hat{\bm{l}}_n$:
\begin{align}
\hat{\bm{l}}_n &= f_{\text{cls}}^n\left(\mathbf{F}'_{m_{n}}\right) \in \mathbb{R}^{T\times 1},
\end{align}
where $f_{\text{cls}}^n(\cdot)$ denotes the classification layer at step $n$.
Subsequently, soft reconstruction converts $\hat{\bm{l}}_n$ into a bit-level rPPG signal $\hat{\bm{y}}_n$:
\begin{align}
\hat{\bm{y}}_n &= \text{Softrecon}(\hat{\bm{l}}_n) \in \mathbb{R}^{T \times 1},
\end{align}
where $\text{Softrecon}(\cdot)$ denotes the soft reconstruction that converts the predicted logits $\hat{\bm{l}}_n$ into the bit-level rPPG signal $\hat{\bm{y}}_n$. Specifically, the soft reconstruction computes the Euclidean distance matrix $\mathbf{D}_n \in \mathbb{R}^{T \times 2^n}$ between the predicted logits $\hat{\bm{l}}_n$ and the codebook values $\mathbf{c}_n$, 
applies a softmax along the codebook dimension to obtain the probability distribution $\mathbf{P}_n = \mathrm{softmax}(-\mathbf{D}_n) \in \mathbb{R}^{T \times 2^n}$, 
and reconstructs the bit-level rPPG signal as $\hat{\bm{y}}_n = \mathbf{P}_n \mathbf{c}_n \in \mathbb{R}^{T \times 1}$.
After the soft reconstruction, the feature projection layer maps $\hat{\bm{y}}_n$ back into the feature space, yielding $\mathbf{F}_{\hat{\bm{y}}_n}$:
\begin{align}
\mathbf{F}_{\hat{\bm{y}}_n} &= f_{\text{proj}}^n(\hat{\bm{y}}_n) \in \mathbb{R}^{T \times C'},
\end{align}
where $f_{\text{proj}}^n(\cdot)$ denotes the feature projection layer at step $n$.
The resulting features $\mathbf{F}_{\hat{\bm{y}}_n}$ are iteratively accumulated with $\mathbf{F}'_{m_{n}}$ over subsequent steps, progressively enriching the temporal representation by combining coarse global rhythms with increasingly fine local variations:
\begin{align}
\mathbf{F}_{m_n} &= \mathbf{F}'_{m_{n}} 
+ \textstyle\sum_{i=1}^{n} \mathbf{F}_{\hat{\bm{y}}_i} \in \mathbb{R}^{T \times C'}.
\end{align}
Overall, this refinement process forms a coarse-to-fine hierarchy: at each step, temporally refined features are produced, bit-level logits are predicted and converted into a bit-level rPPG signal, and the reconstructed signal is projected back and accumulated, enabling the representation to integrate stable global rhythms first and progressively incorporate finer local variations.

Finally, the last refined features $\mathbf{F}_{m_{N-1}}$ are fed into the rPPG estimator, denoted as $f_{\text{est}}(\cdot)$.
The estimator is composed of four sequential 1D convolutional layers with a kernel size of $k=3$, designed to model the temporal characteristics of $\mathbf{F}_{m_{N-1}}$.
The first two convolutional layers are followed by a GELU activation.
The last 1D convolutional layer reduces the feature dimension from $C'$ to 1, producing the $N$-bit logit sequence $\hat{\bm{l}}_N$, which is then converted into the final rPPG signal $\hat{\bm{y}}$ through the soft reconstruction function.

\subsection{Optimization objective for coarse-to-fine rPPG estimation model}
The C2F model aims to progressively refine temporal representations in a coarse-to-fine manner and achieve accurate rPPG signal estimation under multi-bit hierarchical supervision.
To achieve this, two loss components are used: a classification loss $\mathcal{L}^{\text{cls}}_{\text{C2F}}$ and a reconstruction loss $\mathcal{L}^{\text{rec}}_{\text{C2F}}$.

The classification loss $\mathcal{L}^{\text{cls}}_{\text{C2F}}$ guides the learning of each estimated bit-level logit $\hat{\bm{l}}_n$ based on its corresponding quantized pseudo label $\bm{y}_n$.
In this process, a distance-based cross-entropy is introduced to more effectively capture the amplitude-aware relationships among codebook entries.
Specifically, the distance matrix $\mathbf{D}_n$ is computed between $\hat{\bm{l}}_n$ and the codebook values $\mathbf{c}_n$, and a softmax is applied to the negative distances to obtain the probability distribution $\mathbf{P}_n$.
Each timestep is then supervised by a one-hot encoded target $\mathbf{Y}_n \in \mathbb{R}^{T \times 2^n}$ representing the code index of the corresponding pseudo label,
where $\mathbf{Y}_n$ is generated by converting the index vector $\bm{i}_n$ from the LQ module into a one-hot matrix.
The classification loss is then formulated as:
\begin{equation}
\mathcal{L}^{\text{cls}}_{\text{C2F}} =
\frac{\lambda_{\text{ce}}}{N} \sum_{n=1}^{N}
\mathrm{CE}(\mathbf{P}_n, \mathbf{Y}_n).
\label{loss:c2f_ce}
\end{equation}
Unlike conventional categorical cross-entropy that treats codebook indices as independent classes, the distance-based formulation leverages the relative distances between logits and codebook values.
By reflecting the proximity between entries, this formulation enables amplitude-aware supervision, which stabilizes the learning process and encourages smooth signal reconstruction without abrupt amplitude jumps.

Similar to the reconstruction loss used in the LQ module (Eq.~\ref{eq:rec_lq}), the reconstruction loss $\mathcal{L}^{\text{rec}}_{\text{C2F}}$ maintains temporal and spectral consistency between the estimated bit-level rPPG signals $\hat{\bm{y}}_n$ and their corresponding quantized pseudo labels $\bm{y}_n$.
The LQ module employs the reconstruction loss to preserve physiological fidelity by aligning the quantized pseudo labels with the continuous physiological signal during the discretization process.
In contrast, the C2F model adopts the same objective to maintain physiological fidelity by guiding the continuous rPPG estimates to align with the quantized pseudo labels under hierarchical supervision throughout the estimation process.
The reconstruction loss combines a negative Pearson correlation term for temporal alignment and a cross-entropy term in the frequency domain for spectral consistency, defined as:
\begin{align}
\mathcal{L}^{\text{rec}}_{\text{C2F}} =
\sum_{n=1}^{N}
\Big[
\lambda_{\text{time}} \cdot \mathrm{Neg}(\hat{\bm{y}}_n, \bm{y}_n)
+ \lambda_{\text{freq}} \cdot \mathrm{CE}\big(\mathrm{maxIndex}(\mathrm{PSD}(\hat{\bm{y}}_n)), \mathrm{PSD}(\bm{y}_n)\big)
\Big],
\label{loss:c2f_reg}
\end{align}
where the first term promotes temporal alignment between $\hat{\bm{y}}_n$ and $\bm{y}_n$, while the second term enforces spectral consistency in the frequency domain.

Overall, the training objective for the C2F model is defined as:
\begin{equation}
\mathcal{L}_{\text{C2F}} =
\mathcal{L}^{\text{cls}}_{\text{C2F}} +
\mathcal{L}^{\text{rec}}_{\text{C2F}}.
\end{equation}
This objective enables the C2F model to progressively refine temporal representations under hierarchical supervision, achieving robust rPPG estimation.
The relative contributions of each loss term are controlled by the coefficients $\lambda_{\text{ce}}$, $\lambda_{\text{time}}$, and $\lambda_{\text{freq}}$.
The coefficients $\lambda_{\text{time}}$ and $\lambda_{\text{freq}}$ are shared between the LQ module and the C2F model, since both adopt the same principle of preserving temporal and spectral consistency between ground-truth and estimated signals.

\subsection*{Notation summary (Sections~\ref{sec:LQ}--\ref{sec:Opt_C2F})}
For ease of reference and improved readability, Table~\ref{tab:notation_3_2_3_5} provides a summary of the key symbols used in Sections~\ref{sec:LQ}--\ref{sec:Opt_C2F}.

\begin{table}[t!]
\centering
\small
\setlength{\tabcolsep}{6pt}
\renewcommand{\arraystretch}{0.9}
\caption{Summary of key symbols used in Sections~\ref{sec:LQ}--\ref{sec:Opt_C2F}.}
\resizebox{!}{0.73\linewidth}{%
\begin{tabular}{p{0.15\linewidth} p{0.58\linewidth} p{0.25\linewidth}}
\toprule
\textbf{Symbol} & \textbf{Meaning} & \textbf{Shape / Type} \\
\midrule
\multicolumn{3}{l}{\textbf{Stage 1: Label quantization module}}\\
$\bm{y}$ & Ground-truth PPG signal & $\mathbb{R}^{T}$ \\
$\bm{y}_n$ & $n$-bit quantized pseudo label & $\mathbb{R}^{T}$ \\
$n$ & Bit depth index & $n\in\{1,\dots,N\}$ \\
$N$ & Maximum bit depth & scalar \\
$\tilde{\bm{y}}$ & Band-pass filtered PPG signal & $\mathbb{R}^{T}$ \\
$E_n(\cdot)$ & $n$-bit quantization encoder & function \\
$\bm{z}_n$ & Latent output of $E_n$ & $\mathbb{R}^{T}$ \\
$\bm{c}_n$ & $n$-bit codebook vector ($2^n$ scalar entries) & $\mathbb{R}^{2^n}$ \\
$\bm{i}_n$ & Code index vector & $\{1,\dots,2^n\}^{T}$ \\
$\mathcal{L}_{\text{LQ}}$ & Total loss for LQ module & scalar \\
$\mathcal{L}^{\text{rec}}_{\text{LQ}},\mathcal{L}^{\text{feat}}_{\text{LQ}}$ & Reconstruction / feature commitment loss (LQ) & scalar \\
\midrule
\multicolumn{3}{l}{\textbf{Stage 2: Coarse-to-fine rPPG estimation model}}\\
$\mathcal{X}$ & Input facial video & $\mathbb{R}^{C \times T \times H \times W}$ \\
$C,T,H,W$ & Channels, temporal length, height, width & integers \\
$\hat{\bm{y}}$ & Final estimated rPPG signal & $\mathbb{R}^{T \times 1}$ \\
$f_{\text{stem}}(\cdot)$ & Frame stem module & function \\
$\mathbf{F}_{\text{stem}}$ & Stemmed temporal features & $\mathbb{R}^{T \times C'}$ \\
$C'$ & Channel dimension of $\mathbf{F}_{\text{stem}}$ & integer \\
$f_{\text{refiner}}(\cdot)$ & Coarse-to-fine refiner & function \\
$\mathbf{F}_{m_{N-1}}$ & Last refined features (output of $f_{\text{refiner}}$) & $\mathbb{R}^{T \times C'}$ \\
$\mathbf{PE}$ & Learnable positional encoding & $\mathbb{R}^{T \times C'}$ \\
$\mathbf{F}_{m_0}$ & Positional-encoded stem features & $\mathbb{R}^{T \times C'}$ \\
$f_{\text{mamba}}^n(\cdot)$ & Bi-Mamba block at step $n$ & function \\
$\mathbf{F}'_{m_n}$ & Bi-Mamba output at step $n$ & $\mathbb{R}^{T \times C'}$ \\
$f_{\text{cls}}^n(\cdot)$ & Classification layer at step $n$ & function \\
$\hat{\bm{l}}_n$ & Bit-level logits at step $n$ & $\mathbb{R}^{T \times 1}$ \\
$\text{Softrecon}(\cdot)$ & Soft reconstruction (logits $\rightarrow$ signal using $\bm{c}_n$) & function \\
$\mathbf{D}_n,\mathbf{P}_n$ & Distance / probability over codebook & $\mathbb{R}^{T \times 2^n}$ \\
$\hat{\bm{y}}_n$ & Reconstructed bit-level rPPG signal at step $n$ & $\mathbb{R}^{T \times 1}$ \\
$f_{\text{proj}}^n(\cdot)$ & Projection layer at step $n$ & function \\
$\mathbf{F}_{\hat{\bm{y}}_n}$ & Projected features from $\hat{\bm{y}}_n$ at step $n$ & $\mathbb{R}^{T \times C'}$ \\
$\mathbf{F}_{m_n}$ & Accumulated features at step $n$ & $\mathbb{R}^{T \times C'}$ \\
$f_{\text{est}}(\cdot)$ & rPPG estimator & function \\
$\hat{\bm{l}}_N$ & Final $N$-bit logits & $\mathbb{R}^{T \times 1}$ \\
$\mathcal{L}_{\text{C2F}}$ & Total loss for C2F model & scalar \\
$\mathcal{L}^{\text{cls}}_{\text{C2F}},\mathcal{L}^{\text{rec}}_{\text{C2F}}$ & Classification / reconstruction loss (C2F) & scalar \\
\bottomrule
\end{tabular}%
}
\label{tab:notation_3_2_3_5}
\end{table}

\label{sec:Opt_C2F}


\section{Experimental results}
\label{sec4}

\subsection{Datasets}
\label{sec4.1}
Five publicly available datasets (PURE, UBFC, COHFACE, V4V, and MMPD) were used to verify the proposed framework. 
\textbf{PURE}~\citep{stricker2014non} contains 60 one-minute RGB videos from 10 subjects performing six head movements, recorded at 30~fps with a 640 $\times$ 480 resolution under controlled indoor lighting. PPG signals, regarded as ground truth, were acquired using a CMS50E oximeter. 
\textbf{UBFC}~\citep{bobbia2019unsupervised} includes 42 RGB videos from 42 subjects solving arithmetic tasks, each approximately one minute in duration, recorded at 30~fps with a 640 $\times$ 480 resolution under indoor lighting with natural light variations. PPG signals were acquired using a CMS50E oximeter.
\textbf{COHFACE}~\citep{heusch2017reproducible} consists of 160 one-minute RGB videos from 40 subjects, recorded at 20~fps with a 640 $\times$ 480 resolution under both stable and dynamic illumination conditions. PPG signals were captured using a CMS50E oximeter.
\textbf{V4V}~\citep{revanur2021v4v} consists of 1,358 RGB videos from 179 subjects, recorded at 25~fps with a 1280 $\times$ 720 resolution under natural and varying illumination. PPG signals were recorded using a Biopac MP150 system. The V4V dataset includes spontaneous head motion and facial expression changes, and it serves as a realistic benchmark for robust rPPG estimation.
\textbf{MMPD}~\citep{tang2023mmpd} contains 660 one-minute RGB videos from 33 subjects, recorded at 30~fps with a 1280 $\times$ 720 resolution using a Galaxy S22 Ultra. PPG signals were measured using an HKG-07C+ oximeter. The MMPD dataset includes multiple lighting conditions (LED high, LED low, incandescent, and natural light), motion types (stationary, head rotation, talking, and walking), and skin tone groups (Fitzpatrick types 3--6), and it serves as a realistic benchmark for robust rPPG estimation.

\subsection{Evaluation metrics}
\label{sec_4.2}
Four standard metrics commonly adopted in rPPG studies were used to evaluate HR estimation performance at the video level: Mean Absolute Error (MAE), Root Mean Squared Error (RMSE), Mean Absolute Percentage Error (MAPE), and Pearson Correlation Coefficient ($\rho$). These metrics were computed from the estimated and ground-truth HR values to jointly assess the estimation error and the correlation between the estimated and ground-truth HR values. Additionally, we used standard deviation (STD), RMSE, and $\rho$ to evaluate HRV estimation performance in terms of low-frequency power (LF) in normalized units (n.u.), high-frequency power (HF) in normalized units (n.u.), and the ratio of LF to HF power (LF/HF).

\subsection{Experiment settings}
\label{sec_4.3}
Experiments were conducted on a hardware environment equipped with an Intel Core i9-10900K CPU, 64~GB DDR4 RAM, and an NVIDIA RTX A5000 GPU. The proposed LQ-rPPG framework was implemented in PyTorch, and all preprocessing and evaluation were conducted using the rPPG-Toolbox~\citep{liu2023rppg} to ensure consistency with previous studies. Specifically, within the rPPG-Toolbox preprocessing pipeline, input videos were segmented into 160-frame clips. For each clip, the face region was detected in the first frame using an OpenCV Haar Cascade-based face detector, and the resulting bounding box was kept fixed for the remaining frames in the clip. The cropped regions were then resized to $128 \times 128$. In the post-processing stage, the estimated rPPG signal was band-pass filtered using a second-order Butterworth filter with cutoff frequencies of 0.75--2.5 Hz. The power spectral density was computed using the Welch method, and heart rate was obtained by selecting the dominant spectral peak. For all datasets, the videos and the temporally aligned ground-truth PPG signals were processed at each dataset’s original video frame rate.

In our experiments, both Stage 1 (LQ module) and Stage 2 (C2F model) were trained with a batch size of 16. Stage 1 and Stage 2 were trained for 30 and 50 epochs, respectively, and the codebook in Stage 1 was updated exclusively using ground-truth PPG from the training split.
For both stages, the AdamW~\citep{loshchilov2017decoupled} optimizer and the OneCycle learning rate scheduler~\citep{smith2018disciplined} were employed.
The learning rate and weight decay were set to $10^{-3}$ and $0$ for Stage 1, and to $3\times10^{-4}$ and $10^{-5}$ for Stage 2, respectively.
The shared loss coefficients were fixed to $\lambda_{\text{time}}=0.2$ and $\lambda_{\text{freq}}=1.0$, and the maximum bit depth was set to $N=5$.
In Stage 1, the feature commitment coefficient was set to $\lambda_{\text{feat}}=0.5$, and in Stage 2, the coefficient for the distance-based classification loss was set to $\lambda_{\text{ce}}=1.0$. Details on selecting $N$ are provided in Section~\ref{sec:opt_bit_depth}, and details on selecting $\lambda_{\text{time}}$, $\lambda_{\text{freq}}$, and $\lambda_{\text{ce}}$ are provided in Section~\ref{sec:effect_cls_loss}.

\paragraph{Fairness of comparison} For fair comparisons, baseline results reported in Tables~\ref{tab:intra_test_1}--\ref{tab:cross_test_2} were aligned with the same rPPG-Toolbox-based protocol whenever possible. Since rPPG-Toolbox provides a standardized benchmarking framework for rPPG methods, comparisons within the toolbox followed consistent dataset splits, preprocessing, HR estimation, evaluation metrics, backbone implementations, and training configurations. Detailed sources of the baseline results are summarized in~\ref{appendix_sources_baseline}.

\subsection{Intra-dataset testing}
To evaluate the performance of the proposed framework, intra-dataset testing was conducted. In the intra-dataset testing, each model was trained and tested on the same dataset following dataset-specific protocols. Specifically, we adopted the data splits used in previous studies: the protocol of \citet{lu2021dual} for the UBFC and PURE datasets, \citet{heusch2017reproducible} for the COHFACE dataset, \citet{revanur2021v4v} for the V4V dataset, and \citet{zou2025rhythmformer} for the MMPD dataset.

\begin{table}[t]
\caption{Intra-dataset testing results on the PURE, UBFC, and COHFACE datasets. The upper block (Green to POS) includes traditional signal processing methods that do not involve learning-based training. The lower block consists of deep learning-based methods. For deep learning-based methods, training and evaluation are conducted on the same dataset. \textbf{Bold} and \underline{underlined} indicate the best and second-best results, respectively. The symbol $\dagger$ denotes results excerpted from the original papers due to unavailable code and not reproduced under the same rPPG-Toolbox-based protocol.}
\centering
\footnotesize
\resizebox{\textwidth}{!}{
\begin{tabular}{lcccccccccccc}
\toprule
& \multicolumn{3}{c}{PURE} & \multicolumn{3}{c}{UBFC} & \multicolumn{3}{c}{COHFACE} \\
\cmidrule(lr){2-4} \cmidrule(lr){5-7} \cmidrule(lr){8-10}
\multirow{2}{*}{Method} & MAE$\downarrow$         & RMSE$\downarrow$        & \multirow{2}{*}{$\rho$$\uparrow$} & MAE$\downarrow$         & RMSE$\downarrow$        & \multirow{2}{*}{$\rho$$\uparrow$} & MAE$\downarrow$         & RMSE$\downarrow$ & \multirow{2}{*}{$\rho$$\uparrow$} \\
                                         & (bpm) & (bpm) &                                   & (bpm) & (bpm) &                                   & (bpm) & (bpm) \\
\midrule
Green~\citep{verkruysse2008remote} & 4.39 & 11.60 & \textbf{0.99} & 7.50 & 14.14 & 0.62 & 10.94 & 16.72 & - \\
ICA~\citep{poh2010noncontact} & 15.23 & 21.25 & - & 5.17 & 11.76 & 0.65 & 8.89 & 14.55 & 0.42 \\
CHROM~\citep{dehaan2013chrom} & 2.07 & 2.50 & \textbf{0.99} & 2.37 & 4.91 & 0.89 & 7.80 & 12.45 & 0.26 \\
POS~\citep{wang2016algorithm} & 3.14 & 10.57 & 0.95 & 4.05 & 8.75 & 0.78 & - & - & - \\
\midrule
HR-CNN~\citep{vspetlik2018visual} & 1.84 & 2.37 & \underline{0.98} & 4.90 & 5.89 & 0.64 & 8.10 & 10.80 & 0.29 \\
DeepPhys~\citep{chen2018deepphys} & 0.83 & 1.54 & \textbf{0.99} & 6.27 & 10.82 & 0.65 & 8.25 & 14.71 & 0.28 \\
PhysNet~\citep{yu2019remote} & 2.10 & 2.60 & \textbf{0.99} & 2.95 & 3.67 & \underline{0.97} & 5.38 & 10.80 & - \\
TS-CAN~\citep{liu2020multi} & 2.48 & 9.01 & 0.92 & 1.70 & 2.72 & \textbf{0.99} & 4.05 & 11.05 & 0.81\\
Gideon et al.~\citep{gideon2021way} & 2.30 & 2.90 & \textbf{0.99} & 1.85 & 4.28 & 0.93 & 1.50 & 4.60 & 0.90 \\
Dual-GAN~\citep{lu2021dual} & 0.82 & 1.31 & \textbf{0.99} & 0.44 & 0.67 & \textbf{0.99} & - & - & - \\
PhysFormer~\citep{yu2022physformer} & 1.10 & 1.75 & \textbf{0.99} & 0.50 & 0.71 & \textbf{0.99} & 0.84 & 2.67 & \textbf{0.99} \\
EfficientPhys~\citep{liu2023efficientphys} & 1.33 & 5.99 & 0.97 & 1.14 & 1.81 & \textbf{0.99} & 6.31 & 16.63 & 0.61 \\ 
TFA-PFE~\citep{li2023learning} & 1.44 & 2.50 & - & 0.76 & 1.62 & - & - & - & - \\
Li et al.~\citep{li2023contactless} & 0.64 & 1.16 & \textbf{0.99} & 0.48 & 0.64 & \textbf{0.99} & - & - & - \\
Contrast-Phys+~\citep{sun2024contrast} & 0.48 & 0.98 & \textbf{0.99} & 0.21 & 0.80 & \textbf{0.99} & - & - & - \\
RhythmFormer~\citep{zou2025rhythmformer} & 0.27 & 0.47 & \textbf{0.99} & 0.50 & 0.78 & \textbf{0.99} & \underline{0.66} & \underline{1.71} & \textbf{0.99} \\
CodePhys$^\dagger$~\citep{chu2025codephys} & 0.39 & 0.83 & \textbf{0.99} & 0.21 & \textbf{0.26} & \textbf{0.99} & 1.19 & 2.75 & 0.97 \\
RhythmMamba~\citep{zou2025rhythmmamba} & \underline{0.23} & \underline{0.34} & \textbf{0.99} & 0.50 & 0.75 & \textbf{0.99} & 1.16 & 2.59 & \underline{0.98} \\
Style-rPPG$^\dagger$~\citep{liu2025style} & 0.39 & 0.62 & \textbf{0.99} & \underline{0.17} & \underline{0.41} & \textbf{0.99} & 1.12 & 2.91 & \underline{0.98} \\
LQ-rPPG (Ours) & \textbf{0.18} & \textbf{0.28} & \textbf{0.99} & \textbf{0.14} & \textbf{0.26} & \textbf{0.99} & \textbf{0.30} & \textbf{0.59} & \textbf{0.99} \\
\bottomrule
\end{tabular}
}
\label{tab:intra_test_1}
\end{table}

Table~\ref{tab:intra_test_1} presents the intra-dataset results on the PURE, UBFC, and COHFACE datasets.
The upper block includes traditional signal processing-based rPPG methods, whereas the lower block reports learning-based methods. Traditional methods are simple and computationally efficient, but rely on fixed and linear color transformations and hand-crafted assumptions, which limit their representational capacity. As a result, they tend to show lower overall performance and less consistent behavior across datasets compared with learning-based approaches.


For the learning-based methods in the lower block, it is worth noting that the PURE, UBFC, and COHFACE datasets were collected in relatively constrained laboratory environments with limited motion and illumination variation, where the performance of recent methods has nearly saturated. Despite this saturation, the proposed LQ-rPPG achieved favorable results compared with recent methods evaluated under matched settings, including RhythmFormer and RhythmMamba, on all three datasets. These results can be attributed to the fact that, even in constrained settings, ground-truth PPG signals still exhibit minor variability (e.g., subtle amplitude fluctuations and waveform-shape changes), which can destabilize optimization or encourage overfitting to fine-grained label variations. By introducing label quantization, LQ-rPPG reduced sensitivity to such fine-grained label variations and provided more stable supervision, leading to strong performance even under near-saturated conditions.

\begin{table}[t!]
\caption{Intra-dataset testing results on the V4V and MMPD datasets. Each method is trained and evaluated on the same dataset.}
\centering
\footnotesize
\resizebox{0.8\textwidth}{!}{
\begin{tabular}{lccccccccc}
\toprule
& \multicolumn{3}{c}{V4V} & \multicolumn{3}{c}{MMPD} \\
\cmidrule(lr){2-4} \cmidrule(lr){5-7}
\multirow{2}{*}{Method} & MAE$\downarrow$ & RMSE$\downarrow$ & \multirow{2}{*}{$\rho$$\uparrow$} 
& MAE$\downarrow$ & RMSE$\downarrow$ & \multirow{2}{*}{$\rho$$\uparrow$} \\
& (bpm) & (bpm) & & (bpm) & (bpm) \\
\midrule
DeepPhys~\citep{chen2018deepphys} & 10.20 & 13.25 & 0.45 & 22.27 & 28.92 & -0.03 \\
PhysNet~\citep{yu2019remote} & 7.44 & 17.52 & 0.62 & 4.80 & 11.80 & 0.60 \\
TS-CAN~\citep{liu2020multi} & 9.42 & 11.95 & 0.47 & 9.71 & 17.22 & 0.44 \\
PhysFormer~\citep{yu2022physformer} & 7.81 & 11.92 & 0.61 & 11.99 & 18.41 & 0.18 \\
EfficientPhys~\citep{liu2023efficientphys} & 8.24 & 12.55 & 0.52 & 13.47 & 21.32 & 0.21 \\
RhythmFormer~\citep{zou2025rhythmformer} & 3.83 & 9.59 & 0.75 & \underline{3.07} & \underline{6.81} & \underline{0.86} \\
RhythmMamba~\citep{zou2025rhythmmamba} & \underline{1.65} & \underline{4.77} & \underline{0.94} & 3.16 & 7.27 & 0.84 \\
LQ-rPPG (Ours) & \textbf{0.73} & \textbf{2.20} & \textbf{0.98} & \textbf{2.67} & \textbf{6.31} & \textbf{0.87} \\
\bottomrule
\end{tabular}
}
\label{tab:intra_test_2}
\end{table}

To further evaluate the robustness of the proposed framework, we conducted experiments on the V4V and MMPD datasets, which are designed to reflect more realistic conditions involving diverse subjects, motions, and lighting variations. The results are summarized in Table~\ref{tab:intra_test_2}. As shown in the table, even on these challenging datasets, LQ-rPPG achieved strong overall performance compared with prior methods. These results demonstrate the advantage of the proposed framework. By leveraging multi-bit pseudo labels with reduced noise and variability in a coarse-to-fine supervision scheme, the proposed framework is first guided by low-bit pseudo labels to stably learn the global rhythmic structure (i.e., the core heartbeat-related pattern). As a result, it is encouraged to focus on physiologically meaningful rhythms that are consistently shared across samples, rather than being overly driven by detailed waveform changes, leading to robust performance under realistic variations in subjects, motions, and lighting. We further assessed the statistical significance of the intra-dataset results on V4V and MMPD using paired two-sided Wilcoxon signed-rank tests, and the results and discussion are summarized in~\ref{appendix_statistic}.

\begin{figure}[t!]
\centering
\includegraphics[width=0.98\linewidth]{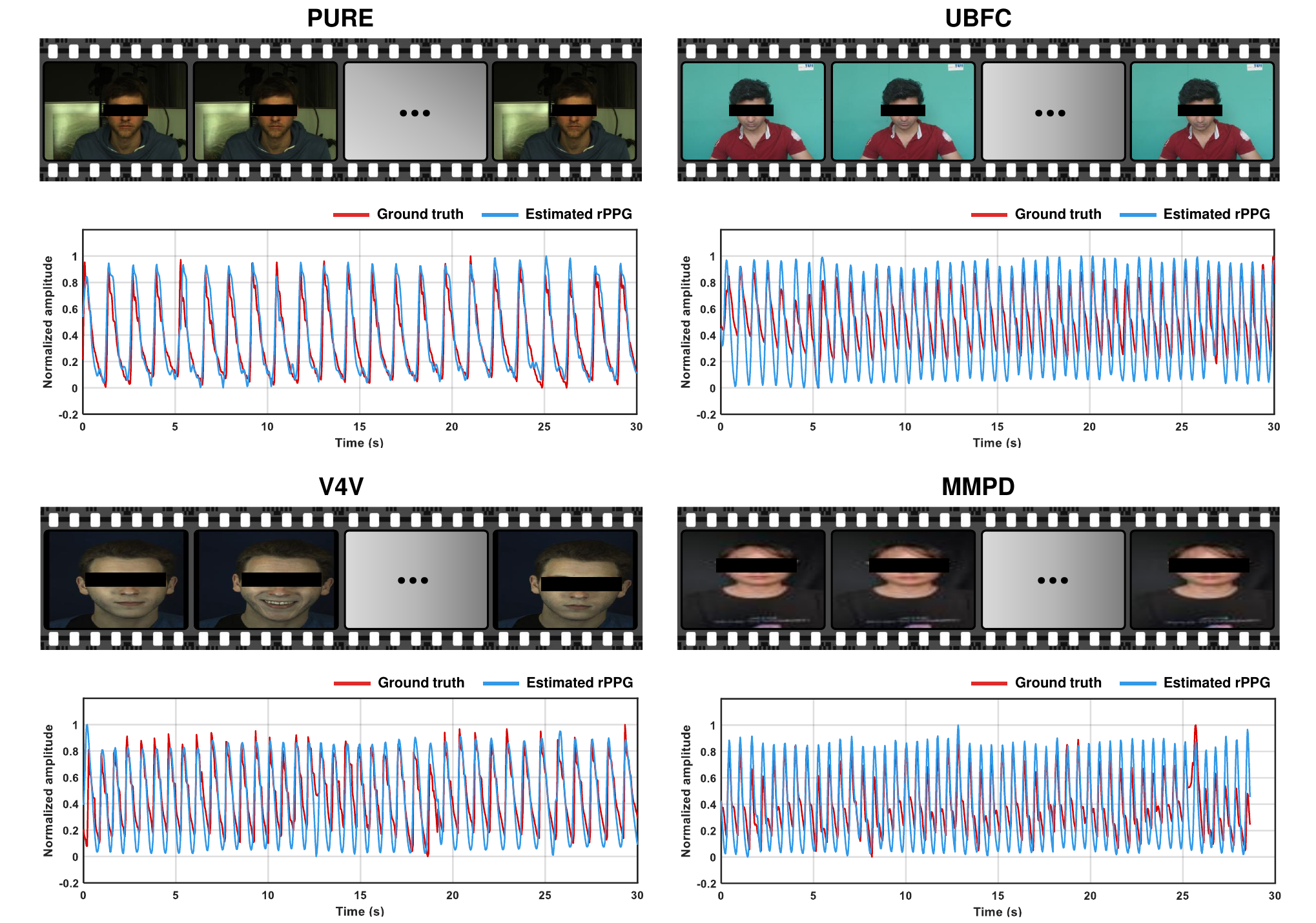}
\caption{Visualization of results on the PURE, UBFC, V4V, and MMPD datasets.}
\label{fig:vis}
\end{figure}

Additionally, we visualized the intra-dataset results to qualitatively evaluate the proposed framework.
Figure~\ref{fig:vis} shows qualitative comparisons between the ground-truth PPG signals and the estimated rPPG signals on the PURE, UBFC, V4V, and MMPD datasets.
As shown in Fig.~\ref{fig:vis}, the proposed LQ-rPPG produces the rPPG signals that are consistent with the ground-truth signals both temporally and periodically across the datasets. In particular, although the ground-truth signal (in red) in the MMPD dataset exhibits noticeable amplitude variations, irregularities in secondary waveform structures, and segment-wise trend changes over certain intervals, the proposed LQ-rPPG maintains stable and consistent predictions (in blue) throughout the entire sequence.
These results further demonstrate that the proposed framework achieves robustness by effectively learning the core physiological patterns rather than following the detailed variability in the ground-truth signals.

Furthermore, to verify that the proposed framework maintains physiologically meaningful beat-to-beat variability beyond HR accuracy, we performed an additional HRV analysis on the UBFC-rPPG dataset. Specifically, following the evaluation protocol of~\citet{sun2022contrast}, we calculated and reported HRV metrics using NeuroKit2~\citep{makowski2021neurokit2}, including LF (n.u.), HF (n.u.), and LF/HF, and evaluated them using STD, RMSE, and $\rho$. We compared the proposed framework with three traditional methods and four deep learning-based methods, as shown in Table~\ref{tab_HRV}. The proposed LQ-rPPG achieved strong overall performance among the compared methods, with particularly strong results on the LF/HF metric, thereby demonstrating that our label-quantized coarse-to-fine supervision preserves physiologically meaningful beat-to-beat variability. Additional HRV results on the more challenging V4V dataset are provided in~\ref{appendix_v4v_hrv}.

\begin{table}[t!]
\caption{HRV results on the UBFC dataset. The upper block (Green to POS) includes traditional signal processing methods that do not involve learning-based training. The lower block consists of deep learning-based methods.}
\centering
\footnotesize
\resizebox{1.0\textwidth}{!}{
\begin{tabular}{lcccccccccccc}
\toprule
& \multicolumn{3}{c}{LF (n.u.)} & \multicolumn{3}{c}{HF (n.u.)} & \multicolumn{3}{c}{LF/HF} \\
\cmidrule(lr){2-4} \cmidrule(lr){5-7} \cmidrule(lr){8-10}
Method & STD$\downarrow$ & RMSE$\downarrow$ & $\rho$$\uparrow$ & STD$\downarrow$ & RMSE$\downarrow$ & $\rho$$\uparrow$ & STD$\downarrow$ & RMSE$\downarrow$ & $\rho$$\uparrow$ \\
\midrule
Green~\citep{verkruysse2008remote} & 0.186 & 0.186 & 0.280 & 0.186 & 0.186 & 0.280 & 0.361 & 0.365 & 0.492 \\
ICA~\citep{poh2010noncontact} & 0.243 & 0.240 & 0.159 & 0.243 & 0.240 & 0.159 & 0.655 & 0.645 & 0.226 \\
POS~\citep{wang2016algorithm} & 0.171 & 0.169 & 0.479 & 0.171 & 0.169 & 0.479 & 0.405 & 0.399 & 0.518 \\
\midrule
Gideon et al.~\citep{gideon2021way} & 0.091 & 0.139 & 0.694 & 0.091 & 0.139 & 0.694 & 0.525 & 0.691 & 0.684 \\
Contrast-Phys~\citep{sun2022contrast} & 0.050 & 0.098 & 0.798 & 0.050 & 0.098 & 0.798 & 0.205 & 0.395 & 0.782 \\
PhysFormer~\citep{yu2022physformer} & \underline{0.030} & \textbf{0.032} & \underline{0.895} & \underline{0.030} & \textbf{0.032} & \underline{0.895} & \underline{0.126} & \underline{0.130} & \underline{0.893} \\
Style-rPPG~\citep{liu2025style} & 0.039 & \underline{0.051} & 0.868 & 0.039 & \underline{0.051} & 0.868 & 0.141 & 0.147 & 0.843 \\
LQ-rPPG (Ours) & \textbf{0.028} & \textbf{0.032} & \textbf{0.902} & \textbf{0.028} & \textbf{0.032} & \textbf{0.902} & \textbf{0.125} & \textbf{0.128} & \textbf{0.899} \\
\bottomrule
\end{tabular}
}
\label{tab_HRV}
\end{table}

\subsection{Cross-dataset testing}
To evaluate the generalization ability of the proposed framework, cross-dataset testing was conducted.
In this experiment, each model was trained on one dataset and tested on another, following the dataset split protocols described in \citet{liu2023rppg} and \citet{zou2025rhythmmamba}.

\begin{table}[t!]
\caption{Cross-dataset testing results on the UBFC $\rightarrow$ PURE and PURE $\rightarrow$ UBFC settings. Each model is trained and validated on the source dataset (left side of ``$\rightarrow$'') and tested on the target dataset (right side of ``$\rightarrow$''). The symbol $\dagger$ denotes results excerpted from the original papers due to unavailable code and not reproduced under the same rPPG-Toolbox-based protocol.}
\centering
\footnotesize
\resizebox{\textwidth}{!}{
\begin{tabular}{lcccccccccccc}
\toprule
& \multicolumn{4}{c}{UBFC $\rightarrow$ PURE} & \multicolumn{4}{c}{PURE $\rightarrow$ UBFC} \\
\cmidrule(lr){2-5} \cmidrule(lr){6-9}
\multirow{2}{*}{Method} 
& MAE$\downarrow$ & RMSE$\downarrow$ & MAPE$\downarrow$ & \multirow{2}{*}{$\rho$$\uparrow$}
& MAE$\downarrow$ & RMSE$\downarrow$ & MAPE$\downarrow$ & \multirow{2}{*}{$\rho$$\uparrow$} \\
& (bpm) & (bpm) & (\%) & & (bpm) & (bpm) & (\%) & \\
\midrule
DeepPhys~\citep{chen2018deepphys} & 5.54 & 18.51 & 5.32 & 0.66 & 1.21 & 2.90 & 1.42 & \textbf{0.99} \\
PhysNet~\citep{yu2019remote} & 8.06 & 19.71 & 13.67 & 0.61 &  0.98 & 2.48 & 1.12 & \textbf{0.99} \\
Meta-rPPG~\citep{lee2020meta} & 4.00 & 5.98 & - & 0.92 & 6.11 & 7.58 & - & 0.66 \\
TS-CAN~\citep{liu2020multi} & 3.69 & 13.80 & 3.39 & 0.82 & 1.30 & 2.87 & 1.50 & \textbf{0.99} \\
Dual-GAN~\citep{lu2021dual} & 1.81 & 2.97 & - & \textbf{0.99} & 0.74 & \underline{1.02} & - & - \\
PhysFormer~\citep{yu2022physformer} & 12.92 & 24.36 & 23.92 & 0.47 & 1.44 & 3.77 & 1.66 & \underline{0.98}  \\
EfficientPhys~\citep{liu2023efficientphys} & 5.47 & 17.04 & 5.40 & 0.71 & 2.07 & 6.32 & 2.10 & 0.94 \\
Spiking-Phys~\citep{liu2025spiking} & 3.83 & - & 5.70 & 0.83 & 2.80 & - & 2.81 & 0.95 \\
RhythmFormer~\citep{zou2025rhythmformer} & 0.97 & 3.36 & \underline{1.60} & \textbf{0.99} & 0.89 & 1.83 & \underline{0.97} & \textbf{0.99}  \\
RhythmMamba~\citep{zou2025rhythmmamba} & 1.98 & 6.51 & 3.59 & \underline{0.96} & 0.95 & 1.83 & 1.04 & \textbf{0.99} \\
Style-rPPG$^\dagger$~\citep{liu2025style} & 0.91 & 1.31 & - & \textbf{0.99} & \underline{0.59} & 1.49 & - & \textbf{0.99} \\
LST-rPPG$^\dagger$~\citep{li2025lst} & \underline{0.49} & \underline{1.19} & - & \textbf{0.99} & 0.98 & 2.32 & - & \textbf{0.99} \\
LQ-rPPG (Ours) & \textbf{0.34} & \textbf{0.65} & \textbf{0.49} & \textbf{0.99} & \textbf{0.46} & \textbf{0.65} & \textbf{0.48} & \textbf{0.99} \\
\bottomrule
\end{tabular}
}
\label{tab:cross_test_1}
\end{table}

Table~\ref{tab:cross_test_1} presents the cross-dataset testing results on the UBFC and PURE datasets. As shown in the table, the proposed LQ-rPPG achieved strong performance across all evaluation metrics, showing relatively small performance degradation compared with the intra-dataset testing results in Table~\ref{tab:intra_test_1}. For instance, when comparing the results on the PURE test set in the intra-dataset testing (trained and tested on PURE) and cross-dataset testing (trained on UBFC and tested on PURE), RhythmFormer degraded from 0.27 to 0.97 MAE (+0.70), and RhythmMamba degraded from 0.23 to 1.98 MAE (+1.75), indicating noticeable degradation under domain shift. In contrast, the proposed LQ-rPPG showed a minor degradation from 0.18 to 0.34 MAE (+0.16). These results indicate that the proposed framework generalizes reliably under domain shift and achieves strong cross-dataset performance. This strong generalization can be attributed to two key design choices in our framework. First, the proposed label quantization reduces label-level domain-dependent variations by suppressing domain-specific differences in the PPG signal (e.g., amplitude scaling and shape changes), thereby making the supervision more consistent across domains. Second, in the coarse-to-fine scheme, low-bit pseudo labels emphasize the global heartbeat rhythm that remains relatively invariant to domain shift, guiding the proposed framework to learn domain-robust rhythmic patterns. Together, these components help mitigate overfitting to domain-specific label characteristics and lead to strong overall performance in cross-dataset evaluation.

\begin{table}[t!]
\caption{Cross-dataset testing results on the UBFC $\rightarrow$ MMPD and PURE $\rightarrow$ MMPD settings. Each model is trained and validated on the source dataset (left side of ``$\rightarrow$'') and tested on the target dataset (right side of ``$\rightarrow$'').}
\centering
\footnotesize
\resizebox{\textwidth}{!}{
\begin{tabular}{lcccccccccccc}
\toprule
& \multicolumn{4}{c}{UBFC $\rightarrow$ MMPD} & \multicolumn{4}{c}{PURE $\rightarrow$ MMPD} \\
\cmidrule(lr){2-5} \cmidrule(lr){6-9}
\multirow{2}{*}{Method} 
& MAE$\downarrow$ & RMSE$\downarrow$ & MAPE$\downarrow$ & \multirow{2}{*}{$\rho$$\uparrow$}
& MAE$\downarrow$ & RMSE$\downarrow$ & MAPE$\downarrow$ & \multirow{2}{*}{$\rho$$\uparrow$} \\
& (bpm) & (bpm) & (\%) & & (bpm) & (bpm) & (\%) & \\
\midrule
DeepPhys~\citep{chen2018deepphys} & 17.50 & 25.00 & 19.27 & 0.06 & 16.92 & 24.61 & 18.54 & 0.05 \\
PhysNet~\citep{yu2019remote} & 9.47 & 16.01 & 11.11 & 0.31 & 13.22 & 19.61 & 14.73 & 0.23 \\
TS-CAN~\citep{liu2020multi} & 14.01 & 21.04 & 15.48 & 0.24 & 13.94 & 21.61 & 15.15 & 0.20 \\
PhysFormer~\citep{yu2022physformer} & 12.10 & 17.79 & 15.41 & 0.17 & 14.57 & 20.71 & 16.73 & 0.15 \\
EfficientPhys~\citep{liu2023efficientphys}  & 13.78 & 22.25 & 15.15 & 0.09 & 14.03 & 21.62 & 15.32 & 0.17 \\
Spiking-Phys~\citep{liu2025spiking} & 14.15 & - & 16.22 & 0.15 & 14.57 & - & 16.55 & 0.14 \\
RhythmFormer~\citep{zou2025rhythmformer} & \underline{9.08} & \underline{15.07} & \underline{11.17} & \textbf{0.53} & \underline{8.98} & \underline{14.85} & \underline{11.11} & \underline{0.51} \\
RhythmMamba~\citep{zou2025rhythmmamba} & 10.63 & 17.14 & 12.14 & 0.34 & 10.44 & 16.70 & 12.25 & 0.36 \\
LQ-rPPG (Ours) & \textbf{8.09} & \textbf{13.94} & \textbf{9.49} & \underline{0.49} & \textbf{7.98} & \textbf{13.35} & \textbf{9.73} & \textbf{0.53} \\
\bottomrule
\end{tabular}
}
\label{tab:cross_test_2}
\end{table}

To further evaluate the generalization ability of the proposed framework under more realistic and challenging conditions, we conducted additional cross-dataset experiments using the MMPD dataset.
Table~\ref{tab:cross_test_2} presents the results, where each model was trained on either the UBFC or PURE dataset and tested on the MMPD dataset.
Since UBFC and PURE are relatively smaller and less diverse, while MMPD contains a wide range of subjects, illumination conditions, and motion variations, this setup provides a highly challenging evaluation of generalization performance. 
For this experiment, we additionally adopted the training protocol of \citet{zou2025rhythmmamba}, which incorporates the data-augmentation strategy proposed by \citet{yu2020autohr}.
As shown in Table~\ref{tab:cross_test_2}, the proposed LQ-rPPG achieved strong overall performance, demonstrating strong generalization under challenging domain shifts.
Specifically, the proposed framework achieved MAE values of 8.09 when trained on UBFC and 7.98 when trained on PURE, which are lower than those of competitive methods such as RhythmFormer (9.08 and 8.98) and RhythmMamba (10.63 and 10.44). These results demonstrate that the proposed LQ-rPPG framework remains effective even under severe domain shifts and limited training conditions, highlighting its potential for deployment in real-world remote physiological monitoring applications. We further assessed the statistical significance of the cross-dataset results on UBFC $\rightarrow$ MMPD and PURE $\rightarrow$ MMPD using paired two-sided Wilcoxon signed-rank tests, and the results and discussion are summarized in~\ref{appendix_statistic}.

\subsection{Computation efficiency analysis}
The previous experimental sections focused on evaluating the accuracy and generalization performance of the proposed framework.
However, for practical deployment, computational efficiency is equally important.
This section quantitatively analyzes the computational efficiency of LQ-rPPG.

\begin{table}[t!]
\caption{Comparison of parameters and computational cost. All models were evaluated under the same input resolution ($128\times128$) and hardware environment (GPU: RTX A5000). Both MACs and throughput were measured per frame, and peak GPU memory was measured for a 160-frame input clip.}
\centering
\footnotesize
\resizebox{0.90\textwidth}{!}{%
\begin{tabular}{lcccc}
\toprule
\multirow{2}{*}{Method} & Param.$\downarrow$ & MACs$\downarrow$ & Throughput$\uparrow$ & Peak GPU memory$\downarrow$ \\
& (M) & (M) & (kfps) & (MB) \\
\midrule
DeepPhys~\citep{chen2018deepphys} & 7.50 & 750.02 & 5.01 & 2687.76 \\
PhysNet~\citep{yu2019remote} & \underline{0.77} & 438.24 & \underline{11.98} & 485.46 \\
TS-CAN~\citep{liu2020multi} & 7.50 & 750.02 & 4.65 & 2786.76 \\
PhysFormer~\citep{yu2022physformer} & 7.38 & 316.29 & 9.47 & 681.45 \\
EfficientPhys~\citep{liu2023efficientphys} & 7.44 & 379.30 & 7.98 & 1747.94 \\
RhythmFormer~\citep{zou2025rhythmformer} & 3.25 & 240.59 & 8.95 & 640.94 \\
CodePhys~\citep{chu2025codephys} & 5.73 & 473.69 & - & - \\
RhythmMamba~\citep{zou2025rhythmmamba} & 1.07 & \underline{80.90} & 7.00 & \underline{449.96} \\
LQ-rPPG (Ours) & \textbf{0.13} & \textbf{57.09} & \textbf{20.35} & \textbf{429.70} \\
\bottomrule
\end{tabular}%
}
\label{tab:computation}
\end{table}

Table~\ref{tab:computation} summarizes the comparison between LQ-rPPG and recent competitive models.
As shown in the table, LQ-rPPG achieved 0.13~M parameters, 57.09~M MACs, and a throughput of 20.35 kfps. Compared with RhythmMamba, a highly competitive lightweight model, the proposed framework reduced the number of parameters by 88\%, computation by 29\%, and increased throughput by 191\%. These direct computational efficiency gains mainly come from the intentionally simplified Mamba-based backbone design. However, the proposed label-quantized coarse-to-fine learning strategy plays a key role in making it practically feasible to adopt such a lightweight design while maintaining strong rPPG estimation performance. Specifically, through the hierarchically guided C2F model, this strategy progressively leverages the multi-bit pseudo labels generated by the LQ module, first learning coarse patterns such as core physiological rhythms under low-bit supervision and then progressively refining finer details under higher-bit supervision. This coarse-to-fine learning process alleviates the learning burden and enables effective training under limited model capacity. This point is further supported by the ablation results on label-quantized coarse-to-fine supervision in Section~\ref{ablation_studies}, particularly in Tables~\ref{tab5_ablation} and~\ref{tab_skip_bit_ablation}.

Furthermore, when considering practical deployment in edge or resource-constrained environments, memory consumption becomes an important consideration. Accordingly, we additionally measured the peak GPU memory usage using a 160-frame input clip. These results are also reported in Table~\ref{tab:computation}. As shown in the table, LQ-rPPG achieved the lowest peak GPU memory usage among the compared methods. These results further indicate that the proposed method is well suited for practical deployment under memory-constrained settings.

Finally, since the proposed framework adopts a two-stage training pipeline, we quantify the additional training-time overhead introduced by Stage~1. Specifically, we compared the training times of Stage~1 and Stage~2 on the UBFC and PURE datasets. Stage~1 required only 1.35 minutes on UBFC and 1.67 minutes on PURE, whereas Stage~2 required 10.94 minutes on UBFC and 15.27 minutes on PURE. As a result, the additional overhead of Stage~1 corresponded to only 12.4\% and 10.9\% of the Stage~2 training time on UBFC and PURE, respectively. Moreover, although retraining Stage~1 on the target dataset is optimal for maximizing pseudo-label fidelity, we found that the LQ module pretrained on a different dataset can still preserve a reasonable level of fidelity (\ref{appendix_lq_transferability}). Therefore, if a pretrained LQ module is available, reusing Stage~1 can be considered depending on the degree of domain shift and the required performance when training on a new domain.

\subsection{Validation of the quantization-based learning strategy}
\label{valid_quantized_learning}
\begin{figure}[t]
\centering
\includegraphics[width=0.98\linewidth]{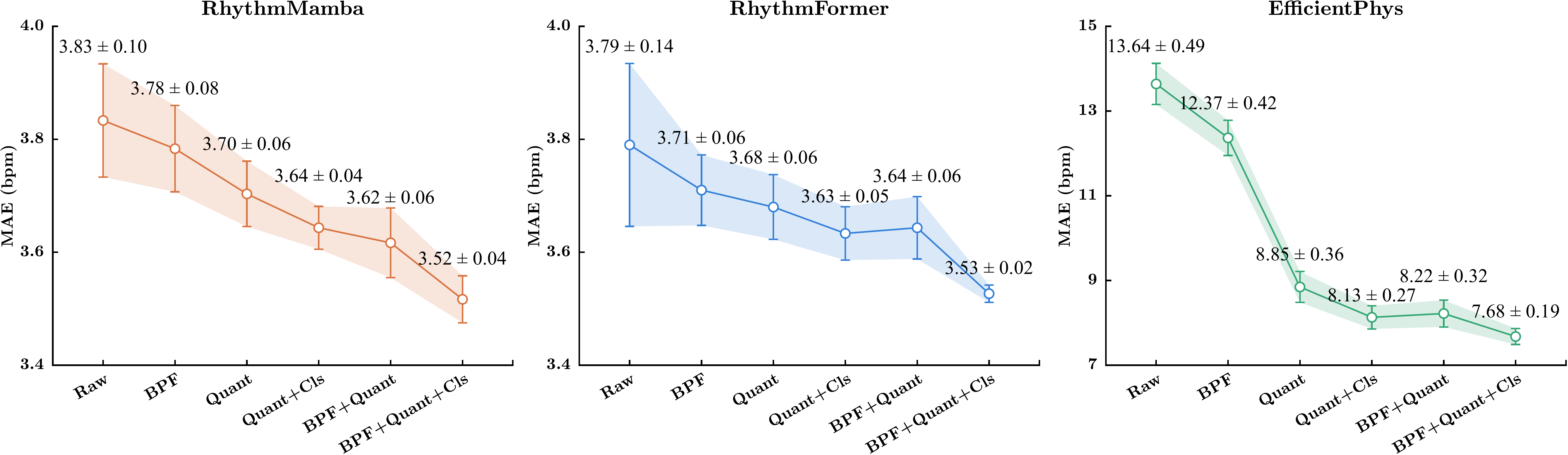}
\caption{Comparison of intra-dataset testing results on the MMPD dataset under six supervision settings: Raw, BPF, Quant, Quant+Cls, BPF+Quant, and BPF+Quant+Cls. Each baseline model was trained with three random seeds (100, 200, and 300), and the mean and standard deviation of MAE are reported.}
\label{fig:comparison_supervision_signal}
\end{figure}

This section demonstrates that label noise and variability degrade the effectiveness of rPPG learning and that quantization-based learning itself can provide a robust solution to this issue.
To verify this hypothesis, three representative models were trained on the MMPD dataset under six different supervision settings at the final output level, without incorporating the coarse-to-fine scheme:
(1) the original PPG labels (Raw),
(2) band-pass-filtered labels (BPF),
(3) quantized labels (Quant),
(4) quantized labels with the classification-based objective (Quant+Cls),
(5) a combination of BPF and quantization (BPF+Quant),
and (6) a combination of BPF and quantization with the classification-based objective (BPF+Quant+Cls), as illustrated in Fig.~\ref{fig:comparison_supervision_signal}.

For a reliable and fair comparison, the six supervision settings were applied to three representative models with different backbone architectures:
the Mamba-based RhythmMamba, the Transformer-based RhythmFormer, and the CNN-based EfficientPhys.
Each experiment was repeated three times with random seeds of 100, 200, and 300, and the mean and standard deviation were reported.

In the Raw setting, the original training pipelines of each baseline model were reproduced without modification.
In the BPF setting, the PPG labels were preprocessed with a band-pass filter to remove non-physiological frequency noise. This setting serves as a reference to quantify the effect of explicit frequency-domain noise removal in the supervision signals.
In the Quant setting, the PPG labels were quantized using the maximum bit level (5-bit), and the models were trained using the quantized labels as supervision. In the Quant+Cls setting, the PPG labels were quantized with the same 5-bit quantization, and the models were trained under quantized supervision, with the proposed classification-based objective additionally applied and rPPG signals generated via soft reconstruction, as in the final output of the C2F model. Finally, in the BPF+Quant and BPF+Quant+Cls settings, the PPG labels were first preprocessed with a band-pass filter and then quantized with the same 5-bit quantization. The former and latter were trained in the same manner as the Quant and Quant+Cls settings, respectively.

As shown in Fig.~\ref{fig:comparison_supervision_signal}, applying BPF improved both accuracy and stability across all models.
The improvement observed in the BPF setting demonstrates that label noise can hinder effective learning, considering that the removal of non-physiological components through band-pass filtering directly enhanced model performance and stability.

The performance gain observed in the Quant setting indicates that quantized supervision serves as an effective solution to mitigate label noise and variability, leading to more stable and generalizable learning. Notably, the Quant setting consistently yielded larger improvements than BPF alone across all models, even without band-pass filtering, suggesting that quantization can effectively cope with label noise and variability inherent in continuous PPG signals. Furthermore, the additional improvement observed in the Quant+Cls setting indicates that the classification-based objective provides complementary benefits on top of quantized supervision.

The performance improvement observed in the BPF+Quant setting over the BPF and Quant settings demonstrates that the two strategies can work complementarily by mitigating noise and variability, and promoting effective learning. BPF primarily suppresses non-physiological frequency components outside the cardiac band, whereas quantization stabilizes supervision by reducing variability and instability in continuous label values. In addition, the BPF+Quant+Cls setting achieved the best overall performance across all models, indicating the benefit of combining the classification-based objective with BPF and quantized supervision.


These results collectively confirm that label noise and variability interfere with effective rPPG learning, and that the quantization-based strategy effectively addresses this limitation. Additional comparisons with alternative noisy-supervision mitigation strategies are provided in~\ref{appendix_alter_strat}.

\subsection{Optimal bit-depth analysis}
\label{sec:opt_bit_depth}
The proposed framework generates multi-level pseudo labels quantized from 1-bit to $N$-bit and performs hierarchical learning based on them. In this structure, the $N$-bit pseudo label used in the final rPPG estimation needs to retain sufficient physiological fidelity to enable precise rPPG estimation.

To determine the optimal value of $N$, we evaluated the MAE between the HR values derived from the quantized pseudo labels and those computed from the original PPG signals across five public datasets (PURE, UBFC, COHFACE, V4V, and MMPD) by progressively increasing the bit level. 
The results are presented in Table~\ref{tab:ablation_n_mae}.
As the bit depth increased from 1 to 4 bits, the MAE gradually decreased, while it nearly converged to zero at 5-bit quantization. No further improvement was observed at 6 bits, indicating that 5-bit pseudo labels are already capable of preserving the essential HR information of the original PPG signals.

\begin{table}[t]
\caption{Comparison of label fidelity across bit levels on five public datasets. Each value denotes the MAE (bpm) between HRs estimated from the quantized pseudo labels and those computed from the PPG signals. The MAE serves as a quantitative measure of label fidelity.}
\centering
\footnotesize
\resizebox{0.6\textwidth}{!}{%
\begin{tabular}{cccccc}
\toprule
 & \multicolumn{5}{c}{Dataset} \\
\cmidrule(lr){2-6}
Bit level ($n$) & PURE & UBFC & COHFACE & V4V & MMPD \\
\midrule
1 & 0.63 & 0.40 & 0.77 & 0.27 & 0.69 \\
2 & 0.32 & 0.14 & 0.11 & 0.15 & 0.39 \\
3 & 0.11 & 0.08 & 0.06 & 0.08 & 0.27 \\
4 & 0.07 & 0.05 & 0.04 & 0.05 & 0.19 \\
5 & \textbf{0.06} & \textbf{0.05} & \textbf{0.03} & \textbf{0.03} & \textbf{0.16} \\
6 & \textbf{0.06} & \textbf{0.05} & \textbf{0.03} & \textbf{0.03} & \textbf{0.16} \\
\bottomrule
\end{tabular}%
}
\label{tab:ablation_n_mae}
\end{table}

\begin{figure}[t]
\centering
\includegraphics[width=0.8\linewidth]{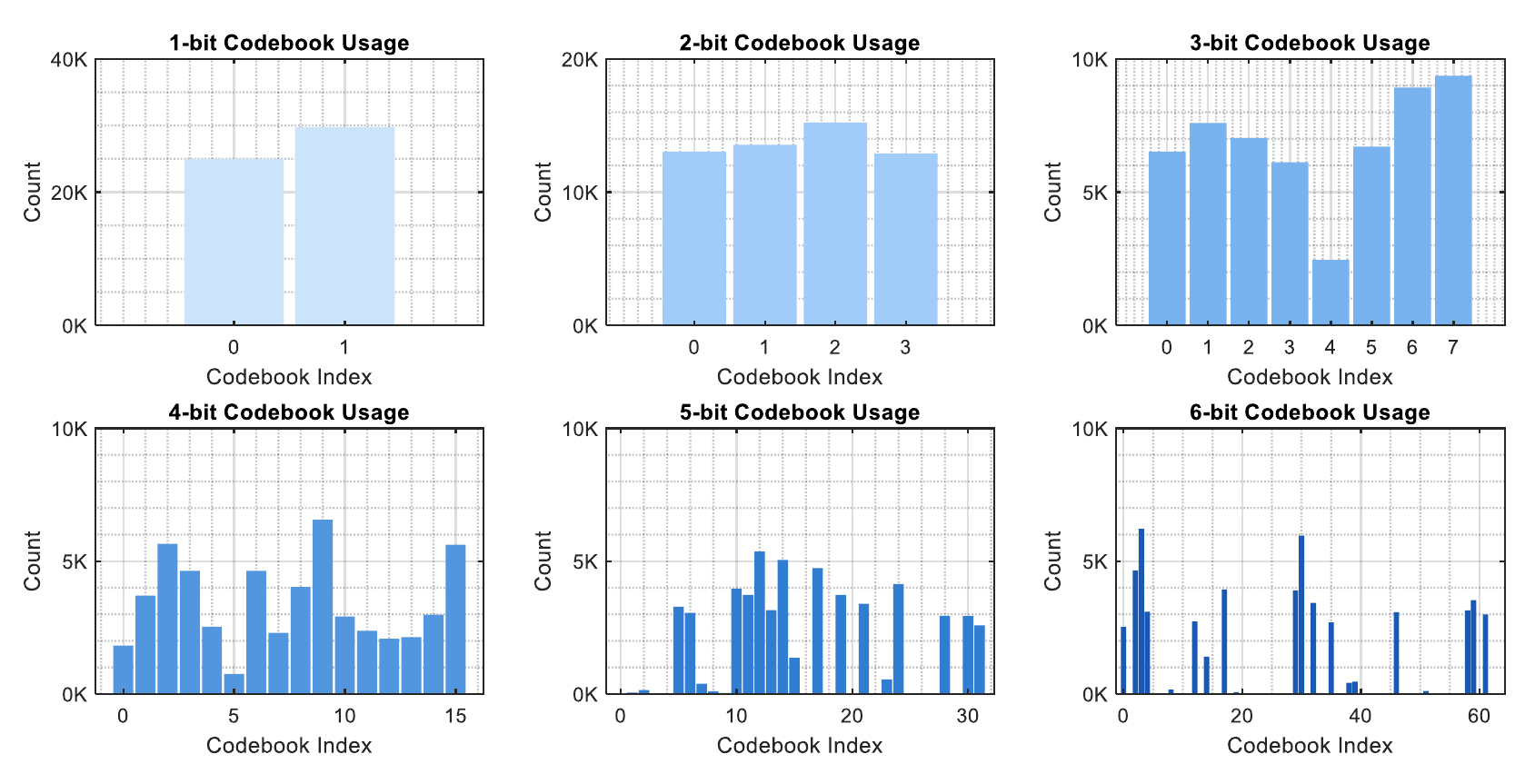}
\caption{Codebook utilization across different bit levels on the UBFC dataset.}
\label{fig:codebook_usage}
\end{figure}

To further verify the optimal bit depth, we analyzed the utilization of codebook entries for each bit level using the UBFC dataset (Fig.~\ref{fig:codebook_usage}). For bit depths from 1 to 4, all codebook entries were used. However, at 5 and 6 bits, several entries were underutilized, indicating saturation in the quantization space. These results suggest that 5 bits are sufficient to represent the original PPG signals.

Based on these findings, we determined that the 5-bit quantization level is the optimal maximum bit depth that preserves label fidelity while maintaining computational efficiency. Accordingly, LQ-rPPG adopts 1- to 5-bit quantization levels in both the label quantization (Stage~1) and coarse-to-fine estimation (Stage~2) to ensure high-fidelity supervision with minimal computational cost.

\subsection{Ablation studies}
\label{ablation_studies}
To verify the contribution of the key design elements in the LQ-rPPG framework, two ablation studies were conducted:
one on the LQ module components,
and the other on the coarse-to-fine learning strategy in the C2F model.

\paragraph{Ablation on the LQ module components}
The LQ module consists of two key components: the dilated convolution block and the Bi-Mamba block.
These components work together to preserve the physiological consistency and temporal structure of the quantized pseudo labels.
The dilated convolution block expands local temporal receptive fields to capture short-range dynamics, while the Bi-Mamba block models long-range temporal dependencies, jointly enhancing the label fidelity of the quantized pseudo labels.

To evaluate the effectiveness of each component, we analyzed the effect of including or excluding each component on the fidelity of the quantized pseudo labels. Specifically, three model variants were tested: (1) using only the dilated convolution block, (2) using only the Bi-Mamba block, and (3) combining both components. The fidelity of the quantized pseudo labels was assessed by measuring the MAE between the HR values derived from the quantized pseudo labels and those computed from the ground-truth PPG signals across five public datasets (PURE, UBFC, COHFACE, V4V, and MMPD).

The results are summarized in Table~\ref{tab:ablation_conv_component}. Across all datasets, the combination of the two components consistently achieved the lowest MAE, indicating that the two components complement each other and jointly improve the physiological fidelity of the quantized pseudo labels.

\begin{table}[t!]
\caption{Comparison of the effects of LQ module components (dilated conv and Bi-Mamba) on the fidelity of the quantized pseudo labels across bit levels and five public datasets. Each value denotes the MAE (bpm) between HRs estimated from the quantized pseudo labels and those computed from the PPG signals. The MAE serves as a quantitative measure of label fidelity.}
\centering
\footnotesize
\renewcommand{\arraystretch}{0.9}
\resizebox{0.7\textwidth}{!}{%
\begin{tabular}{lcc|ccccc}
\toprule
\multirow{3}{*}{Dataset} 
& \multicolumn{2}{c|}{Ablation setting} 
& \multicolumn{5}{c}{Bit level ($n$)} \\
\cmidrule(lr){2-3} \cmidrule(lr){4-8}
 & Dilated conv & Bi-Mamba 
 & 1 & 2 & 3 & 4 & 5 \\
\midrule
\multirow{3}{*}{PURE}
 & \checkmark  &  & 0.72 & 0.52 & 0.22 & 0.09 & 0.08 \\
 &  & \checkmark  & 0.66 & 0.42 & 0.20 & 0.09 & 0.07 \\
 & \checkmark & \checkmark 
 & \textbf{0.63} & \textbf{0.32} & \textbf{0.11} & \textbf{0.07} & \textbf{0.06} \\
\midrule
\multirow{3}{*}{UBFC}
 & \checkmark  &  & 0.50 & 0.22 & 0.19 & 0.14 & 0.08 \\
 &  & \checkmark  & 0.47 & 0.21 & 0.11 & 0.09 & 0.07 \\
 & \checkmark & \checkmark 
 & \textbf{0.40} & \textbf{0.14} & \textbf{0.08} & \textbf{0.05} & \textbf{0.05} \\
\midrule
\multirow{3}{*}{COHFACE}
 & \checkmark & & 0.87 & 0.22 & 0.17 & 0.11 & 0.08   \\
 &  & \checkmark & 0.88 & 0.21 & 0.15 & 0.08 & 0.06 \\
 & \checkmark & \checkmark 
 & \textbf{0.77} & \textbf{0.11} & \textbf{0.06} & \textbf{0.04} & \textbf{0.03} \\
\midrule
\multirow{3}{*}{V4V}
 & \checkmark &  & 0.51 & 0.24 & 0.17 & 0.12 & 0.09 \\
 &  & \checkmark & 0.51 & 0.23 & 0.16 & 0.10 & 0.08 \\
 & \checkmark & \checkmark 
 & \textbf{0.27} & \textbf{0.15} & \textbf{0.08} & \textbf{0.05} & \textbf{0.03} \\
\midrule
\multirow{3}{*}{MMPD}
 & \checkmark &  & 0.78 & 0.44 & 0.35 & 0.22 & 0.21 \\
 &  & \checkmark & 0.77 & 0.45 & 0.33 & 0.22 & 0.18 \\
 & \checkmark & \checkmark 
 & \textbf{0.69} & \textbf{0.39} & \textbf{0.27} & \textbf{0.19} & \textbf{0.16} \\
\bottomrule
\end{tabular}%
}
\label{tab:ablation_conv_component}
\end{table}

\paragraph{Ablation on the coarse-to-fine learning}
The C2F model learns robust and generalized representations by training on the multi-bit quantized pseudo labels in a coarse-to-fine manner. To validate this learning strategy, we evaluated the model on the challenging MMPD dataset by progressively incorporating supervision from higher-bit to lower-bit levels.

Table~\ref{tab5_ablation} summarizes the results.
The performance consistently improved as lower-bit pseudo labels were progressively added, demonstrating the effectiveness of hierarchical supervision.
In particular, clear performance gains were observed when 1-bit and 2-bit labels were included, suggesting that coarse supervision from the low-bit labels helps the model learn stable and physiologically meaningful representations.
When all 1--5 bit levels were jointly used, the model achieved the best overall performance.
These results indicate that the hierarchical coarse-to-fine learning strategy enables the model to first learn stable representations from low-bit supervision and then refine them with higher-bit labels, resulting in improved overall performance and robustness.

\begin{table}[t!]
\caption{Performance on the MMPD dataset with progressively added multi-bit pseudo label supervision.}
\centering
\footnotesize
\renewcommand{\arraystretch}{0.9}
\setlength{\tabcolsep}{2.5mm}
\begin{tabular}{ccccc|ccc}
\toprule
\multicolumn{5}{c|}{Bit level ($n$)} 
& \multirow{2}{*}[-0.5ex]{\makecell[c]{MAE$\downarrow$ \\ (bpm)}} 
& \multirow{2}{*}[-0.5ex]{\makecell[c]{RMSE$\downarrow$ \\ (bpm)}} 
& \multirow{2}{*}[-0.5ex]{$\rho$$\uparrow$} \\
\cmidrule(lr){1-5}
1 & 2 & 3 & 4 & 5 & & & \\
\midrule
& & & & \checkmark & 3.68 & 7.72 & 0.82 \\
& & & \checkmark & \checkmark & 3.57 & 7.61 & 0.82 \\
& & \checkmark & \checkmark & \checkmark & 3.45 & 7.25 & 0.83 \\
& \checkmark & \checkmark & \checkmark & \checkmark & 3.22 & 6.87 & 0.84 \\
\checkmark & \checkmark & \checkmark & \checkmark & \checkmark & \textbf{2.67} & \textbf{6.31} & \textbf{0.87} \\
\bottomrule
\end{tabular}
\label{tab5_ablation}
\end{table}

Additionally, to clarify the contribution of each bit level, we conducted leave-one-bit-out ablation studies on the MMPD dataset by excluding supervision at each bit level, while keeping the model unchanged. As shown in Table~\ref{tab_skip_bit_ablation}, excluding 1-bit supervision led to the largest performance degradation, suggesting the importance of coarse low-bit supervision. At the same time, excluding the 2--4-bit supervision levels also consistently degraded performance. This consistent degradation suggests that the intermediate bit levels act as transitional supervision, forming a bridge between coarse structural guidance from low-bit labels and fine-grained information from higher-bit labels, and thus play a meaningful role in the coarse-to-fine learning process.

\begin{table}[t!]
\caption{Performance on the MMPD dataset with leave-one-bit-out multi-bit pseudo label supervision.}
\centering
\footnotesize
\renewcommand{\arraystretch}{0.9}
\setlength{\tabcolsep}{2.5mm}
\begin{tabular}{ccccc|ccc}
\toprule
\multicolumn{5}{c|}{Bit level ($n$)} 
& \multirow{2}{*}[-0.5ex]{\makecell[c]{MAE$\downarrow$ \\ (bpm)}} 
& \multirow{2}{*}[-0.5ex]{\makecell[c]{RMSE$\downarrow$ \\ (bpm)}} 
& \multirow{2}{*}[-0.5ex]{$\rho$$\uparrow$} \\
\cmidrule(lr){1-5}
1 & 2 & 3 & 4 & 5 & & & \\
\midrule
\checkmark & \checkmark & \checkmark & \checkmark & \checkmark & \textbf{2.67} & \textbf{6.31} & \textbf{0.87} \\
 & \checkmark & \checkmark & \checkmark & \checkmark & 3.22 & 6.87 & 0.84 \\
\checkmark &  & \checkmark & \checkmark & \checkmark & 3.12 & 6.78 & 0.85 \\
\checkmark & \checkmark &  & \checkmark & \checkmark & 2.87 & 6.81 & 0.85 \\
\checkmark & \checkmark & \checkmark &  & \checkmark & 2.89 & 6.63 & 0.85 \\
\bottomrule
\end{tabular}
\label{tab_skip_bit_ablation}
\end{table}

We also conducted a sensitivity analysis to quantitatively assess the impact of the maximum bit depth $N$ by varying $N$ from 2 to 6, considering both the intra-dataset evaluation on MMPD and the cross-dataset setting (UBFC $\rightarrow$ MMPD). As shown in Table~\ref{tab:ablation_max_bit_depth}, performance improved substantially as $N$ increased from 2 to 5, whereas further increasing $N$ to 6 yielded only a limited improvement in both settings. These results can be interpreted based on Table~\ref{tab:ablation_n_mae} and Fig.~\ref{fig:codebook_usage}. Table~\ref{tab:ablation_n_mae} shows that pseudo-label fidelity generally improved as the bit level increased, but the improvement saturated once $N \geq 5$, with very similar fidelity at 5-bit and 6-bit levels. In addition, Fig.~\ref{fig:codebook_usage} indicates that the effective codebook usage saturated when $N \geq 5$, where several codebook entries were underutilized, suggesting that 5-bit quantization was already sufficient to represent the original PPG signals. Taken together, increasing the bit depth beyond 5 provides only limited additional supervisory information during training, which explains why increasing $N$ from 5 to 6 leads to only marginal performance gains in downstream C2F model training.

\begin{table}[t!]
\caption{Sensitivity analysis of the maximum bit depth ($N$) under the intra-dataset setting on MMPD and the cross-dataset setting (UBFC $\rightarrow$ MMPD).}
\centering
\footnotesize
\renewcommand{\arraystretch}{0.9}
\setlength{\tabcolsep}{2.6mm}
\begin{tabular}{ccccccc}
\toprule
\multirow{3}{*}[-0.5ex]{\makecell[c]{Maximum \\ bit depth ($N$)}} 
& \multicolumn{3}{c}{MMPD} 
& \multicolumn{3}{c}{UBFC $\rightarrow$ MMPD} \\
\cmidrule(lr){2-4}\cmidrule(lr){5-7}
& \makecell[c]{MAE$\downarrow$ \\ (bpm)}
& \makecell[c]{RMSE$\downarrow$ \\ (bpm)}
& $\rho\uparrow$
& \makecell[c]{MAE$\downarrow$ \\ (bpm)}
& \makecell[c]{RMSE$\downarrow$ \\ (bpm)}
& $\rho\uparrow$ \\
\midrule
2 & 4.65 & 10.27 & 0.62 & 15.50 & 22.42 & 0.14 \\
3 & 3.46 &  7.50 & 0.83 & 12.14 & 18.23 & 0.18 \\
4 & 2.88 &  6.74 & 0.85 &  9.87 & 15.18 & 0.41 \\
5 & \textbf{2.67} & 6.31 & \textbf{0.87} & \textbf{8.09} & \textbf{13.94} & \textbf{0.49} \\
6 & 2.71 & \textbf{6.30} & \textbf{0.87} & 8.21 & 14.12 & \textbf{0.49} \\
\bottomrule
\end{tabular}
\label{tab:ablation_max_bit_depth}
\end{table}

\subsection{Effect of the classification loss}
\label{sec:effect_cls_loss}
The rPPG estimation in the C2F model is guided by the two loss functions defined in Eq.~\ref{loss:c2f_ce} and Eq.~\ref{loss:c2f_reg}.
Among them, Eq.~\ref{loss:c2f_reg} adopts the regression-based loss structure used in~\cite{zou2025rhythmformer, zou2025rhythmmamba},
which combines the negative Pearson correlation loss and the PSD-based cross-entropy loss weighted by $\lambda_\text{time}$ and $\lambda_\text{freq}$, respectively.
This loss configuration has been validated in \cite{zou2025rhythmformer, zou2025rhythmmamba}, and we followed the same setup by fixing $\lambda_\text{time}=0.2$ and $\lambda_\text{freq}=1.0$.

In this study, we additionally introduce a classification loss, implemented as a distance-based cross-entropy formulation defined in Eq.~\ref{loss:c2f_ce}, to support quantization-based learning within the C2F model.
The classification loss enables the model to learn structured representations within the quantized label space, complementing the regression-based objectives that preserve the physiological continuity and precision of reconstructed rPPG signals.

\begin{figure}
\centering
\includegraphics[width=0.60\linewidth]{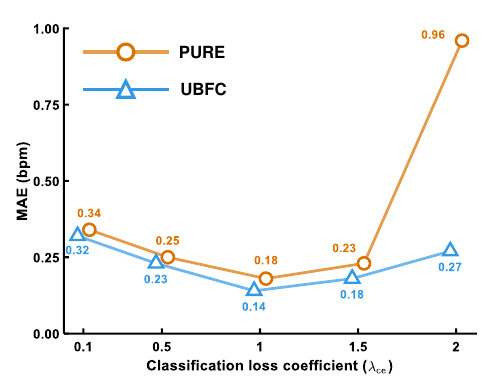}
\caption{Effect of the classification loss coefficient $\lambda_\text{ce}$ on the MAE between the estimated and ground-truth HRs across the PURE and UBFC datasets.}
\label{CE}
\end{figure}

To evaluate the effect of this additional loss and its interaction with the regression terms, we conducted experiments by varying $\lambda_\text{ce}$ across \{0.1, 0.5, 1.0, 1.5, 2.0\}, while keeping $\lambda_\text{time}$ and $\lambda_\text{freq}$ fixed.
The experiments were performed on the PURE and UBFC datasets, and model performance was measured by the MAE between the estimated and ground-truth HRs, as summarized in Figure~\ref{CE}.
As shown in the figure, $\lambda_\text{ce}=1.0$ achieved the lowest MAE on both datasets, demonstrating a well-balanced trade-off between classification and regression supervision.
When $\lambda_\text{ce}$ was too small, the influence of classification supervision was weakened, leading to underutilization of structural relationships within the quantized label space and resulting in unstable soft reconstructions.
Conversely, when $\lambda_\text{ce}$ was excessively large, the model became overly biased toward discrete codebook indices, producing rigid reconstructions and reducing physiological smoothness.
Therefore, $\lambda_\text{ce}=1.0$ provides a good balance between discrete classification and continuous regression learning, resulting in stable convergence and superior overall performance across datasets.

\subsection{Effect of end-to-end joint training}
To investigate the effect of joint optimization, we additionally evaluated an end-to-end variant of our framework on the V4V and MMPD datasets, where the LQ module and the C2F model were trained jointly within a single training pipeline. Unlike the two-stage setting, pseudo labels were generated on-the-fly by the LQ module during C2F training, and both modules were updated simultaneously using the combined loss. We used the same loss functions and hyperparameter settings as in the two-stage training. The total number of training epochs was set to 50, matching the Stage 2 training length, and no additional tuning was applied to the joint training setting.

\begin{figure}[t!]
\centering
\includegraphics[width=1.0\linewidth]{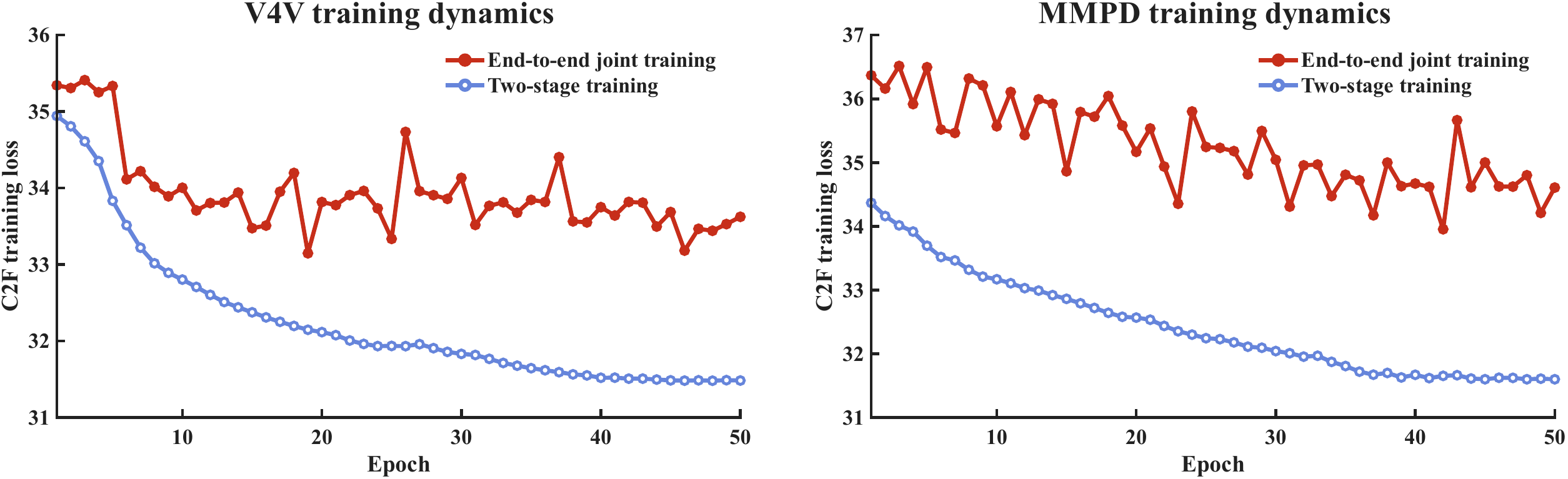}
\caption{Comparison of C2F model training loss curves under the two-stage setting and the end-to-end setting on V4V and MMPD.}
\label{Training_curve}
\end{figure}

The results are summarized in Table~\ref{tab:end-to-end}. Under the current experimental setup, the end-to-end setting showed worse performance than the two-stage setting across both datasets. In addition, as shown in Fig.~\ref{Training_curve}, the end-to-end setting exhibited more oscillatory training dynamics and remained at a higher training loss level than the two-stage setting. These observations suggest that jointly optimizing the LQ module and the C2F model may make the intermediate supervision less stable than in the two-stage setting. Because the pseudo labels are generated on-the-fly while the LQ module is being updated, they may not provide stable supervision throughout training, which can make the C2F objective harder to optimize. This effect may be particularly unfavorable for coarse-to-fine learning, which relies on stable intermediate supervision across bit levels.

\begin{table}[t!]
\caption{Effect of end-to-end joint training on the V4V and MMPD datasets.}
\centering
\footnotesize
\resizebox{0.75\textwidth}{!}{
\begin{tabular}{lcccccc}
\toprule
& \multicolumn{3}{c}{V4V} & \multicolumn{3}{c}{MMPD} \\
\cmidrule(lr){2-4} \cmidrule(lr){5-7}
\multirow{2}{*}{Method} & MAE$\downarrow$ & RMSE$\downarrow$ & \multirow{2}{*}{$\rho$$\uparrow$} 
& MAE$\downarrow$ & RMSE$\downarrow$ & \multirow{2}{*}{$\rho$$\uparrow$} \\
& (bpm) & (bpm) & & (bpm) & (bpm) \\
\midrule
LQ-rPPG (end-to-end) & 3.81 & 9.65 & 0.76 & 7.88 & 17.96 & 0.31 \\
LQ-rPPG (two-stage) & \textbf{0.73} & \textbf{2.20} & \textbf{0.98} & \textbf{2.67} & \textbf{6.31} & \textbf{0.87} \\
\bottomrule
\end{tabular}
}
\label{tab:end-to-end}
\end{table}

\subsection{Effect of ROI re-detection frequency}

In our preprocessing, the face region was detected in the first frame of each clip and kept fixed for the entire clip. Since ROI settings can influence input quality under motion, we conducted an auxiliary analysis to examine the sensitivity of LQ-rPPG to ROI re-detection frequency in cross-dataset scenarios.

Specifically, we evaluated LQ-rPPG on UBFC → MMPD and PURE → MMPD using three ROI settings: per-frame ROI re-detection (LQ-rPPG-PF), 1-second ROI re-detection (LQ-rPPG-1s), and the default fixed ROI used in our method (LQ-rPPG). As shown in Table~\ref{tab:ROI_setting}, the default fixed ROI achieved the best results among the evaluated settings in both scenarios, suggesting that, for LQ-rPPG, simply increasing the ROI re-detection frequency did not provide additional benefit. However, this analysis is limited to the evaluated settings and does not constitute a comprehensive comparison of broader ROI strategies. More advanced ROI strategies may still lead to further performance improvements.

\begin{table}[t!]
\caption{Comparison of LQ-rPPG under different ROI re-detection frequencies on UBFC $\rightarrow$ MMPD and PURE $\rightarrow$ MMPD. LQ-rPPG-PF and LQ-rPPG-1s denote per-frame and 1-second ROI re-detection, respectively, while LQ-rPPG uses the default fixed ROI.}
\centering
\footnotesize
\resizebox{\textwidth}{!}{
\begin{tabular}{lcccccccccccc}
\toprule
& \multicolumn{4}{c}{UBFC $\rightarrow$ MMPD} & \multicolumn{4}{c}{PURE $\rightarrow$ MMPD} \\
\cmidrule(lr){2-5} \cmidrule(lr){6-9}
\multirow{2}{*}{Method} 
& MAE$\downarrow$ & RMSE$\downarrow$ & MAPE$\downarrow$ & \multirow{2}{*}{$\rho$$\uparrow$}
& MAE$\downarrow$ & RMSE$\downarrow$ & MAPE$\downarrow$ & \multirow{2}{*}{$\rho$$\uparrow$} \\
& (bpm) & (bpm) & (\%) & & (bpm) & (bpm) & (\%) & \\
\midrule
LQ-rPPG-PF & 10.73 & 16.89 & 11.42 & 0.46 & 11.48 & 17.33 & 13.95 & 0.41 \\
LQ-rPPG-1s & 8.81 & 15.42 & 10.44 & 0.47 & 8.92 & 15.23 & 11.21 & 0.51 \\
LQ-rPPG & \textbf{8.09} & \textbf{13.94} & \textbf{9.49} & \textbf{0.49} & \textbf{7.98} & \textbf{13.35} & \textbf{9.73} & \textbf{0.53} \\
\bottomrule
\end{tabular}
}
\label{tab:ROI_setting}
\end{table}

\subsection{Comparison with non-learnable quantization}
\label{sec_nonlearnable}
To further verify the necessity of the proposed learnable LQ module in Stage~1, we constructed a non-learnable uniform quantization setting as a counterpart to Stage~1 and compared the two on the MMPD dataset. First, we evaluated how well each quantization scheme preserves HR information in the ground-truth PPG signal. Specifically, pseudo labels were generated using either uniform quantization applied after band-pass filtering (for a fair comparison with the proposed LQ module) or the proposed LQ module, and we computed the MAE between the HR estimated from each pseudo label and the ground-truth HR. As summarized in Table~\ref{tab:uniform_vs_lq_mmpd}, uniform quantization generally preserved HR information less effectively than the proposed LQ module. In addition, the HR MAE under uniform quantization did not decrease monotonically with increasing bit-width (e.g., the MAE increased to 1.14 at 4-bit), suggesting unstable HR preservation. This can be attributed to the fact that uniform quantization discretized amplitudes with fixed bins without explicitly preserving HR-relevant time–frequency structure, whereas the proposed LQ module was trained to preserve such structure through the reconstruction loss in Eq.~(\ref{eq:rec_lq}), which enforces temporal and spectral consistency.

This label-level trend was also consistent with downstream rPPG estimation results. As shown in Table~\ref{tab:uniform_vs_lq_mmpd}, training the C2F model with pseudo labels generated by the proposed LQ module achieved substantially better performance on MMPD than training with uniform-quantized pseudo labels. These results demonstrate that the proposed learnable LQ module is important for generating HR-consistent pseudo labels that provide effective supervision for coarse-to-fine rPPG learning, and that non-learnable uniform quantization is not a sufficient substitute.

\begin{table}[t]
\caption{Comparison between non-learnable uniform quantization and the proposed learnable LQ module on MMPD. Left: HR MAE between pseudo labels and ground-truth PPG. Right: downstream rPPG performance using the C2F model.}
\centering
\footnotesize
\resizebox{\textwidth}{!}{%
\begin{tabular}{ccccccccc}
\toprule
 & \multicolumn{5}{c}{HR MAE$\downarrow$ (pseudo label vs. GT) } & \multicolumn{3}{c}{Downstream performance} \\
\cmidrule(lr){2-6}\cmidrule(lr){7-9}
Method & 1-bit & 2-bit & 3-bit & 4-bit & 5-bit & MAE $\downarrow$ & RMSE $\downarrow$ & $\rho$ $\uparrow$ \\
\midrule
Uniform quantization & 2.18 & 0.96 & 0.94 & 1.14 & 0.83 & 4.84 & 12.31 & 0.61 \\
Proposed LQ module   & \textbf{0.69} & \textbf{0.39} & \textbf{0.27} & \textbf{0.19} & \textbf{0.16} & \textbf{2.67} & \textbf{6.31}  & \textbf{0.87} \\
\bottomrule
\end{tabular}%
}
\label{tab:uniform_vs_lq_mmpd}
\end{table}

\section{Limitations and failure cases}
\label{limit_failure}
The proposed framework is designed to stabilize supervision under noisy and variable supervisory PPG signals, allowing it to learn heartbeat-related physiological features more consistently across diverse conditions and thereby improving robustness and generalization. However, the proposed framework does not include a dedicated mechanism to explicitly restore or correct severely corrupted inputs (e.g., strong motion-induced artifacts), nor does it incorporate sufficiently sophisticated spatiotemporal modeling to learn motion-robust representations at the feature level and suppress such artifacts. Consequently, the framework can still be limited in scenarios involving extreme motion, where facial optical signals are heavily disrupted.

\begin{figure}[t!]
\centering
\includegraphics[width=1.0\linewidth]{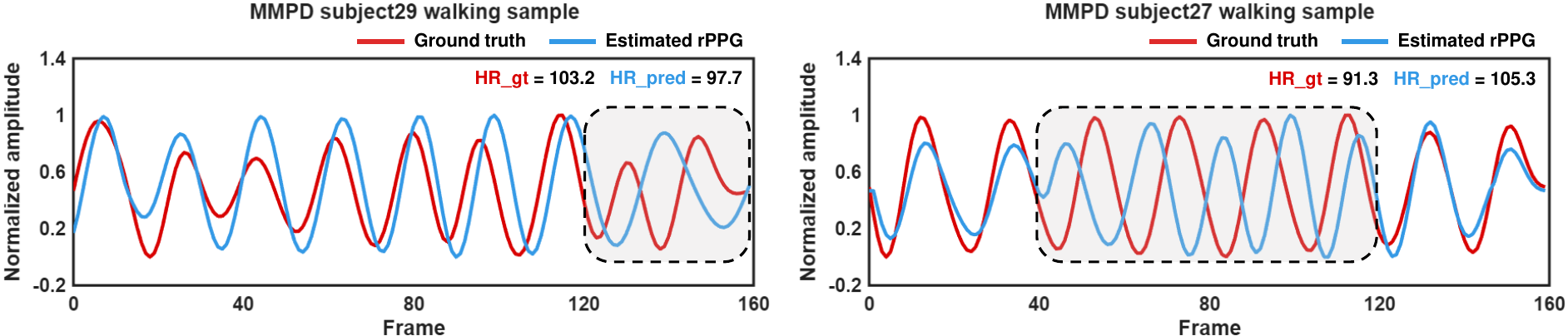}
\caption{Qualitative failure cases in the MMPD walking scenario. The estimated rPPG (blue) and the ground-truth PPG (red) are both band-pass filtered to emphasize the dominant cardiac component for clearer rhythm comparison.}
\label{failure_cases}
\end{figure}

These limitations were occasionally observed in the walking scenario in the MMPD dataset. The walking scenario is particularly challenging because it involves simultaneous camera motion and facial motion, which can easily increase motion artifacts and substantially disrupt facial optical signals. Figure~\ref{failure_cases} illustrates representative failure cases during abrupt motion bursts, where the estimated rPPG signal temporarily lost its periodicity within the dashed gray boxes (left: frames 120--160; right: frames 40--120). These observations suggest that, while the proposed framework improves robustness and generalization overall, extreme motion conditions may still call for complementary components that more directly mitigate severe input corruption or provide more sophisticated spatiotemporal feature modeling.

\section{Conclusion}
\label{sec5}
This study proposes LQ-rPPG, a label-quantized coarse-to-fine learning framework for robust rPPG estimation.
To address the challenges posed by label noise and variability in PPG signals used as ground truth for rPPG model training, the framework transforms PPG signals into multi-bit quantized pseudo labels and leverages them to guide rPPG estimation in a coarse-to-fine manner.
This label-centric strategy enables LQ-rPPG to integrate the stability of low-bit supervision with the fidelity of high-bit guidance, allowing the framework to learn structured physiological representations and achieve robust rPPG estimation across diverse conditions.
Extensive experiments across multiple public datasets demonstrate that LQ-rPPG achieves strong performance in both intra-dataset and cross-dataset settings.
Furthermore, the framework exhibits strong computational efficiency, requiring only 0.13~M parameters and 57.09~M MACs, and achieving a throughput of 20.35~kfps.
Overall, LQ-rPPG establishes a robust and efficient paradigm for rPPG estimation, showing strong potential for real-world deployment.

This study further suggests that, in rPPG learning, performance and generalization are influenced not only by model design and representation learning, but also by the quality and consistency of supervisory PPG labels. Therefore, alongside efforts to improve model design and representation learning, a label-centric perspective on how supervisory signals are defined and handled should also be considered in rPPG learning.

\paragraph{Future work}
Since this study primarily focuses on a label-centric learning approach, the framework is intentionally kept simple and efficient. 
Such simplicity reduces computational cost but limits the capacity to capture complex visual variations and global texture dependencies. 
This limitation may be more pronounced at scale and in real-world deployment, where input conditions are more diverse and temporally variable. Future work will therefore improve the representational capacity of the framework with spatiotemporal modeling that captures spatial and temporal dynamics more effectively. We will then assess scalability on larger datasets and longer sequences, and evaluate deployment performance under continuous streaming and online inference using latency, memory footprint, and prediction stability.

Additionally, the proposed label-quantized coarse-to-fine supervision has the potential to generalize to other continuous-signal learning tasks where supervision is noisy or variable. Exploring and validating such applicability beyond rPPG remains a direction for future research.

\paragraph{Supervision realism and deployment considerations}
The proposed LQ-rPPG is a supervised framework that relies on contact-based PPG signals during training and is not intended to replace unsupervised, self-supervised, or weakly supervised approaches. Rather than targeting label scarcity in terms of quantity, LQ-rPPG focuses on a practical constraint in real-world data collection: although contact-based PPG signals can often be obtained, their quality is frequently unstable due to motion artifacts, subject-dependent physiological differences, and domain shifts across sensors, acquisition setups, and environments. To account for this, the framework is designed to mitigate the impact of label noise and variability, enabling robust rPPG estimation even when the available supervisory signals are imperfect. Consequently, when contact-based PPG signals are extremely scarce or difficult to acquire, unsupervised, self-supervised, or weakly supervised methods may be more suitable. In contrast, when contact-based PPG signals are available but noisy or unstable in realistic training settings, the proposed supervision can be a practical and effective choice.

\section*{Code availability}
The code is available at the public repository~\citep{lq_rppg_code}.




\appendix
\renewcommand{\thetable}{A.\arabic{table}}
\renewcommand{\thefigure}{A.\arabic{figure}}
\renewcommand{\theHtable}{A.\arabic{table}}
\renewcommand{\theHfigure}{A.\arabic{figure}}
\setcounter{table}{0} 
\setcounter{figure}{0} 

\section{Quantitative analysis of PPG label quality}
\label{appendix_quant_label_quality}

To provide quantitative evidence on the quality of contact-based PPG labels, we computed a frequency-domain signal-to-noise ratio (SNR) for the ground-truth PPG signals from the datasets used in this work. Each PPG sequence was normalized using clip-wise z-score normalization, after which the power spectral density was estimated. The SNR was defined as the ratio between the spectral power within the physiological heart-rate band (0.7--2.5 Hz) and the spectral power outside this band. Table~\ref{tab:SNR_label} reports the mean and standard deviation of the resulting SNR values for each dataset. Overall, the results show that the quality of contact-based PPG labels is not uniform across datasets, with several datasets exhibiting low and highly variable SNR, indicating that contact-based PPG signals used as ground truth are often not ideal.


\section{Sources of baseline results and protocol alignment}
\label{appendix_sources_baseline}
This appendix briefly summarizes the sources of baseline results reported in Tables~\ref{tab:intra_test_1}--\ref{tab:cross_test_2} and clarifies the protocol alignment used for fair comparisons. Whenever possible, we conducted comparisons under the same rPPG-Toolbox-based protocol, either by using baseline results reported under the protocol or by reproducing baseline results with public code. Specifically, for baselines whose results were already reported under the same rPPG-Toolbox-based protocol in RhythmMamba and RhythmFormer, we directly cited those paper-reported results for comparison. For the remaining methods, we reproduced results under the same protocol when public code was available; otherwise, we used the baseline results reported in the original papers.

\begin{table}[t!]
\caption{Frequency-domain SNR (dB) of ground-truth PPG labels.}
\centering
\footnotesize
\resizebox{0.6\textwidth}{!}{
\begin{tabular}{lc}
\toprule
Dataset & SNR (dB, mean $\pm$ std) \\
\midrule
PURE~\citep{stricker2014non} & 6.66 $\pm$ 2.34 \\
UBFC~\citep{bobbia2019unsupervised} & 4.22 $\pm$ 2.11 \\
COHFACE~\citep{heusch2017reproducible} & 6.45 $\pm$ 3.63\\
V4V~\citep{revanur2021v4v} & 1.01 $\pm$ 3.24 \\
MMPD~\citep{tang2023mmpd} & 0.81 $\pm$ 3.65\\
\bottomrule
\end{tabular}
}
\label{tab:SNR_label}
\end{table}

\begin{itemize}
    \item \textbf{Table~\ref{tab:intra_test_1} (PURE/UBFC/COHFACE, intra-dataset)}: Baseline results on PURE/UBFC were taken from RhythmMamba, while baseline results on COHFACE were taken from RhythmFormer, both under the same rPPG-Toolbox-based protocol. Since RhythmMamba does not report results on COHFACE, we additionally ran the publicly available RhythmMamba code on COHFACE to obtain the corresponding results. CodePhys and Style-rPPG were included using the results reported in their papers due to unavailable code.
    
    \item \textbf{Table~\ref{tab:intra_test_2} (V4V/MMPD, intra-dataset)}: Baseline results on V4V were reproduced under the same protocol using publicly available code, and baseline results on MMPD were taken from RhythmMamba.

    \item \textbf{Table~\ref{tab_HRV} (HRV analysis on UBFC)}: This experiment followed the same evaluation setting adopted in Style-rPPG (i.e., the \citet{sun2022contrast} setting), and baseline results were taken from Style-rPPG.

    \item \textbf{Table~\ref{tab:cross_test_1} (UBFC$\rightarrow$PURE/PURE$\rightarrow$UBFC, cross-dataset)}: Cross-dataset baseline results were taken from RhythmMamba, while Style-rPPG and LST-rPPG were included using the results reported in their papers due to unavailable code.
    
    \item \textbf{Table~\ref{tab:cross_test_2} (UBFC$\rightarrow$MMPD/PURE$\rightarrow$MMPD, cross-dataset)}: Baseline results were taken from RhythmMamba.
\end{itemize}

\section{Cross-dataset transferability of the LQ module}
\label{appendix_lq_transferability}
To examine whether the Stage~1 LQ module generalizes under cross-domain shifts, we conducted a cross-dataset fidelity analysis. Specifically, we trained the LQ module only on the UBFC training split and then froze its parameters and codebook. We subsequently applied this UBFC-trained LQ module to the training splits of other datasets (PURE, COHFACE, V4V, and MMPD) to generate 1--6-bit quantized pseudo labels. To ensure consistency across datasets with different sampling rates, we resampled the ground-truth PPG signals to a unified temporal resolution of 30 Hz prior to quantization. For each bit level, we measured label fidelity using the MAE (bpm) between HRs estimated from the quantized pseudo labels and those computed from the ground-truth PPG signals, as summarized in Table~\ref{tab:lq_transfer_fidelity}. 

\begin{table}[t]
\caption{Cross-dataset pseudo-label fidelity across bit levels using the LQ module trained on UBFC. Each value denotes the MAE (bpm) between HRs estimated from the quantized pseudo labels and those computed from the PPG signals. The MAE serves as a quantitative measure of label fidelity.}
\centering
\footnotesize
\resizebox{0.6\textwidth}{!}{%
\begin{tabular}{ccccc}
\toprule
 & \multicolumn{4}{c}{Target dataset (training split)} \\
\cmidrule(lr){2-5}
Bit level ($n$) & PURE & COHFACE & V4V & MMPD \\
\midrule
1 & 0.64 & 1.35 & 1.08 & 0.77 \\
2 & 0.63 & 0.40 & 0.40 & 0.44 \\
3 & 0.19 & 0.18 & 0.37 & 0.27 \\
4 & 0.08 & 0.09 & 0.15 & 0.20 \\
5 & 0.06 & 0.08 & 0.13 & 0.20 \\
6 & 0.06 & 0.09 & 0.13 & 0.20 \\
\bottomrule
\end{tabular}%
}
\label{tab:lq_transfer_fidelity}
\end{table}

Overall, the results show that pseudo-label fidelity can be maintained at a reasonable level even under domain shifts. Across all target datasets, the MAE typically decreased as the bit level increased and converged to a low-error regime at higher bit levels. In particular, from 2-bit onward, the MAE consistently remained below 1 bpm across all datasets, suggesting that the pretrained LQ module preserves quantization fidelity across domains to a meaningful extent. Notably, while the contact-based PPG signals in UBFC were measured using a CMS50E oximeter, those in V4V and MMPD were collected using different devices (Biopac MP150 and an HKG-07C+ oximeter, respectively); despite these differences, the UBFC-pretrained LQ module maintained reasonable pseudo-label fidelity across datasets. This behavior is plausible because, despite substantial differences in sensors and recording conditions, PPG signals share common physiological structures, such as cardiac-induced periodicity and basic waveform patterns. Taken together, while retraining Stage~1 on the target domain is preferable for maximizing pseudo-label fidelity, our findings also suggest that a pretrained LQ module can be conditionally reused without retraining depending on the degree of domain shift and the required performance.



\section{Statistical significance analysis}
\label{appendix_statistic}
To verify whether the performance gains of the proposed framework are statistically significant, we conducted paired two-sided Wilcoxon signed-rank tests for the intra-dataset results on V4V and MMPD in Table~\ref{tab:intra_test_2} and the cross-dataset results on UBFC $\rightarrow$ MMPD and PURE $\rightarrow$ MMPD in Table~\ref{tab:cross_test_2}. For each setting, we selected the baseline that achieved the second-best performance on the corresponding dataset and evaluated the statistical significance of the improvement over that baseline. For example, on V4V in Table~\ref{tab:intra_test_2}, we compared the proposed framework with RhythmMamba, and on MMPD, we compared the proposed framework with RhythmFormer.

For the Wilcoxon signed-rank test, we computed the absolute HR error $AE$ for each sample as the absolute difference between the predicted HR and the ground-truth HR, and defined the paired difference as $\Delta = AE_{\text{baseline}} - AE_{\text{LQ-rPPG}}$. The results are summarized in Table~\ref{tab:wilcoxon_sig}, where $N$ denotes the number of samples used in the test. Win, Tie, and Loss denote the numbers of samples with $\Delta > 0$, $\Delta = 0$, and $\Delta < 0$, respectively, providing an intuitive summary of how often LQ-rPPG yields lower, equal, or higher errors than the baseline. $\mathrm{median}(\Delta)$ denotes the median of $\Delta$ over all samples, summarizing the typical magnitude of the paired difference.

\begin{table}[t!]
\caption{Paired two-sided Wilcoxon signed-rank test results under intra-dataset and cross-dataset settings.}
\centering
\footnotesize
\resizebox{0.85\textwidth}{!}{
\begin{tabular}{l l c c c c}
\toprule
Dataset setting & Baseline & $N$ & Win/Tie/Loss & $\mathrm{median}(\Delta)$ & p-value \\
\midrule
\multicolumn{6}{l}{\textbf{Intra-dataset}} \\
\addlinespace[0.8mm]
V4V  & RhythmMamba  & 357 & 220/52/85 & 0.3301 & $2.11\times10^{-19}$ \\
MMPD & RhythmFormer & 140 & 76/29/35  & 0.5401 & $1.15\times10^{-3}$ \\
\addlinespace[1.2mm]
\midrule
\multicolumn{6}{l}{\textbf{Cross-dataset}} \\
\addlinespace[0.8mm]
UBFC $\rightarrow$ MMPD & RhythmFormer & 660 & 361/80/219 & 0.5401 & $4.38\times10^{-13}$ \\
PURE $\rightarrow$ MMPD & RhythmFormer & 660 & 379/42/239 & 1.0801 & $3.56\times10^{-10}$ \\
\bottomrule
\end{tabular}
}
\label{tab:wilcoxon_sig}
\end{table}

As shown in Table~\ref{tab:wilcoxon_sig}, the p-values are below the 0.05 significance level across all settings, indicating that the performance gains of the proposed framework are statistically significant. Moreover, the Win counts are larger than the Loss counts in all settings, suggesting that the improvements are not confined to a small subset of samples but are consistently observed across many samples.

\section{Statistical uncertainty analysis}
\label{appendix_variability}
To strengthen statistical reporting beyond significance testing, we additionally report the standard error (SE) of LQ-rPPG across all intra- and cross-dataset settings. The standard error is an indicator of the statistical accuracy of an estimate such as the mean, and is equal to the standard deviation of the estimate’s theoretical sampling distribution under repeated sampling. It also accounts for the number of samples used in measurement, which is particularly useful for rPPG datasets where the number of test samples can vary substantially across datasets. Following the definition in rPPG-Toolbox~\citep{liu2023rppg}, we compute the SE for MAE, RMSE, and $\rho$ and report the results in the form of metric $\pm$ SE, as summarized in Table~\ref{tab:lq_se_all}.

\begin{table}[t!]
\caption{Intra- and cross-dataset results of LQ-rPPG reported as metric $\pm$ SE.}
\centering
\footnotesize
\resizebox{0.80\textwidth}{!}{
\begin{tabular}{lccc}
\toprule
Dataset setting & MAE$\downarrow$ (bpm) & RMSE$\downarrow$ (bpm) & $\rho$$\uparrow$ \\
\midrule
\multicolumn{4}{l}{\textbf{Intra-dataset}} \\
\addlinespace[1mm]
PURE    & 0.18 $\pm$ 0.07 & 0.28 $\pm$ 0.07 & 0.99 $\pm$ 0.00 \\
UBFC    & 0.14 $\pm$ 0.07 & 0.26 $\pm$ 0.04 & 0.99 $\pm$ 0.01 \\
COHFACE & 0.30 $\pm$ 0.06 & 0.59 $\pm$ 0.17 & 0.99 $\pm$ 0.01 \\
V4V     & 0.73 $\pm$ 0.11 & 2.20 $\pm$ 1.60 & 0.98 $\pm$ 0.01 \\
MMPD    & 2.67 $\pm$ 0.48 & 6.31 $\pm$ 12.27 & 0.87 $\pm$ 0.04 \\
\addlinespace[1.5mm]
\midrule
\multicolumn{4}{l}{\textbf{Cross-dataset}} \\
\addlinespace[1mm]
UBFC $\rightarrow$ PURE & 0.34 $\pm$ 0.07 & 0.65 $\pm$ 0.15 & 0.99 $\pm$ 0.00 \\
PURE $\rightarrow$ UBFC & 0.46 $\pm$ 0.07 & 0.65 $\pm$ 0.11 & 0.99 $\pm$ 0.01 \\
UBFC $\rightarrow$ MMPD & 8.09 $\pm$ 0.44 & 13.94 $\pm$ 17.87 & 0.49 $\pm$ 0.03 \\
PURE $\rightarrow$ MMPD & 7.98 $\pm$ 0.42 & 13.35 $\pm$ 15.25 & 0.53 $\pm$ 0.03 \\
\bottomrule
\end{tabular}
}
\label{tab:lq_se_all}
\end{table}

\section{Analysis of noisy-supervision mitigation strategies}
\label{appendix_alter_strat}

This appendix compares the proposed quantization-based learning strategy with representative alternatives from the perspective of robust learning under noisy supervision. Representative approaches for handling noisy or unstable supervision include robust objectives that reduce the influence of outliers, label-level processing that smooths, denoises, or transforms the target labels, and uncertainty-aware learning that adjusts the learning contribution of highly noisy segments based on estimated uncertainty. We instantiate these approaches with Huber-based robust regression, moving-average label smoothing, and uncertainty-aware learning implemented with a Gaussian negative log-likelihood loss. Meanwhile, our quantization-based learning strategy builds on label-level processing by denoising and transforming the label via band-pass filtering and quantization, but it can be viewed as a supervision reformulation beyond conventional label-level processing, as it couples a classification-based objective with a structured label representation.

Experiments followed the protocol used in Section~\ref{valid_quantized_learning}, with RhythmFormer and RhythmMamba as the evaluation backbones. Table~\ref{tab:alt_strategies} reports intra-dataset MAE on MMPD (mean $\pm$ std over three seeds) under five settings. Raw and BPF+Quant correspond to the original and quantization-based supervision settings in Section~\ref{valid_quantized_learning}, respectively. For the alternative strategies, Huber augmented the original loss with an auxiliary Huber (SmoothL1) loss; MA applied a moving-average filter with a window size of 5 to the original PPG labels while keeping the loss composition unchanged; and Gaussian NLL added a linear layer to jointly predict the mean and uncertainty and optimized a Gaussian negative log-likelihood loss in addition to the original loss.

\begin{table}[t!]
\caption{Intra-dataset MAE (mean $\pm$ std) on MMPD under different noisy-supervision learning strategies.}
\centering
\footnotesize
\resizebox{\textwidth}{!}{
\begin{tabular}{lccccc}
\toprule
Backbone & Raw & Huber & MA & Gaussian NLL & BPF+Quant \\
\midrule
RhythmMamba  & 3.83 $\pm$ 0.08 & 4.78 $\pm$ 0.18 & 5.43 $\pm$ 0.37 & 3.76 $\pm$ 0.06 & 3.52 $\pm$ 0.04 \\
RhythmFormer & 3.79 $\pm$ 0.14 & 4.53 $\pm$ 0.18 & 5.72 $\pm$ 0.44 & 3.74 $\pm$ 0.11 & 3.53 $\pm$ 0.02 \\
\bottomrule
\end{tabular}
}
\label{tab:alt_strategies}
\end{table}

As shown in Table~\ref{tab:alt_strategies}, Huber and MA degraded performance relative to Raw for both backbones, Gaussian NLL yielded a modest improvement, and BPF+Quant showed the most consistent gains and stability. Huber reduced the impact of isolated outliers, but it did not address segment-level amplitude and waveform distortions common in rPPG labels; thus, models still tended to follow these time-domain variations, which limited spectral-peak formation and led to no performance gain. MA suppressed high-frequency fluctuations, but it also attenuated true cardiac components or distorted waveform morphology, weakening heart-rate rhythm information and resulting in performance degradation compared with Raw. Gaussian NLL improved optimization stability by predicting uncertainty and downweighting highly noisy segments, but it did not explicitly remove non-physiological components or structural variability in the labels, which limited the achievable gains. In contrast, BPF+Quant filtered out non-physiological components and reduced label variability via quantization, then leveraged a classification loss to learn from the resulting structured supervision, yielding the most consistent improvements and stable training. Furthermore, quantization can be extended to coarse-to-fine supervision via multi-bit label generation, which may further improve performance.

\section{Additional HRV analysis on the V4V dataset}
\label{appendix_v4v_hrv}

To further examine whether the proposed framework preserves beat-to-beat variability under more challenging conditions, we additionally conducted HRV analysis on the V4V dataset. We compared LQ-rPPG with PhysFormer, which achieved the second-best performance in Table~\ref{tab_HRV}. As shown in Table~\ref{tab_HRV_v4v}, LQ-rPPG achieved better results than PhysFormer across LF, HF, and LF/HF metrics, suggesting that the proposed framework can preserve beat-to-beat variability even under noisier conditions.

\begin{table}[!t]
\caption{HRV results of PhysFormer and LQ-rPPG on the V4V dataset.}
\centering
\footnotesize
\resizebox{1.0\textwidth}{!}{
\begin{tabular}{lcccccccccccc}
\toprule
& \multicolumn{3}{c}{LF (n.u.)} & \multicolumn{3}{c}{HF (n.u.)} & \multicolumn{3}{c}{LF/HF} \\
\cmidrule(lr){2-4} \cmidrule(lr){5-7} \cmidrule(lr){8-10}
Method & STD$\downarrow$ & RMSE$\downarrow$ & $\rho$$\uparrow$ & STD$\downarrow$ & RMSE$\downarrow$ & $\rho$$\uparrow$ & STD$\downarrow$ & RMSE$\downarrow$ & $\rho$$\uparrow$ \\
\midrule
PhysFormer~\citep{yu2022physformer} & 0.152 & 0.155 & 0.636 & 0.152 & 0.155 & 0.636 & 1.866 & 1.885 & 0.465 \\
LQ-rPPG (Ours) & \textbf{0.138} & \textbf{0.140} & \textbf{0.681} & \textbf{0.138} & \textbf{0.140} & \textbf{0.681} & \textbf{1.632} & \textbf{1.654} & \textbf{0.539} \\
\bottomrule
\end{tabular}
}
\label{tab_HRV_v4v}
\end{table}

\clearpage
\bibliography{reference}

\end{document}